\documentclass{article}

\usepackage[preprint,nonatbib]{neurips_2023}

\usepackage[utf8]{inputenc} % allow utf-8 input
\usepackage[T1]{fontenc}    % use 8-bit T1 fonts
\usepackage{url}            % simple URL typesetting
\usepackage{booktabs}       % professional-quality tables
\usepackage{amsfonts}       % blackboard math symbols
\usepackage{nicefrac}       % compact symbols for 1/2, etc.
\usepackage{microtype}      % microtypography
\usepackage[T1]{fontenc} % https://www.overleaf.com/learn/latex/International_language_support#Font_encoding
\usepackage{tgtermes}
\usepackage[utf8]{inputenc}
\usepackage{amssymb,latexsym}
\usepackage[nointegrals]{wasysym}

\usepackage[parfill]{parskip} % space between paragraphs
\usepackage{afterpage}
\usepackage{enumitem}
\usepackage{framed}
\usepackage{xspace}

\usepackage{setspace}
\usepackage{stackengine}

{\endMakeFramed}

\usepackage{lineno}

\usepackage{ragged2e}

\newcounter{parcount}

\usepackage[usenames,dvipsnames]{xcolor}
\definecolor{shadecolor}{gray}{0.9}

\usepackage{siunitx}  % \SI{1e-4}{\metre}
\usepackage{amsmath}
\usepackage{euscript}
\usepackage{amsthm}
\newtheorem{thm}{Theorem}
\newtheorem{lemma}{Lemma}

\newtheorem{prop}{Proposition}

\newtheorem{asum}{Assumption}

\usepackage{booktabs,array}

\usepackage{graphicx}
\usepackage[labelfont=bf]{caption}
\usepackage{subcaption}
\graphicspath{{./},{../figures/},{./figures/}}
\usepackage{float}
\usepackage{wrapfig}

\usepackage[square,numbers]{natbib} % numbers is more compact, 
\usepackage[colorlinks,linktoc=all]{hyperref}
\usepackage[all]{hypcap}
\hypersetup{citecolor=MidnightBlue}
\hypersetup{linkcolor=MidnightBlue}
\hypersetup{urlcolor=MidnightBlue}

\usepackage[nameinlink]{cleveref}
\crefname{equation}{Eq}{Eq}
\Crefname{equation}{Eq}{Eq}
\crefname{algorithm}{Alg}{Alg}
\Crefname{algorithm}{Alg}{Alg}
\crefname{table}{Tab}{Tab}
\Crefname{table}{Tab}{Tab}
\crefname{figure}{Fig}{Fig}
\Crefname{figure}{Fig}{Fig}
\crefname{appendix}{App}{App}
\Crefname{appendix}{App}{App}
\crefname{section}{Sec}{Sec}
\Crefname{section}{Sec}{Sec}
\crefname{thm}{Thm}{Thm}
\Crefname{thm}{Thm}{Thm}
\crefname{prop}{Prop}{Prop}
\Crefname{prop}{Prop}{Prop}
\crefname{condition}{Condition}{Conditions}
\Crefname{condition}{Condition}{Conditions}
\crefname{asum}{Assumption}{Assumption}
\Crefname{asum}{Assumption}{Assumption}

\usepackage
[acronym,nowarn,section,nonumberlist]{glossaries}
\glsdisablehyper{}

\usepackage{algorithm}
\usepackage[noend]{algpseudocode}

\usepackage{url}
\usepackage{datetime}
\usepackage{environ}

\newcommand{\atphantom}{\vphantom{${}^2$}}
\newcommand{\AProcedure}[2]{\Procedure{\smash{#1}}{\smash{#2}}}
\newcommand{\AComment}[1]{\Comment{\smash{#1}}}
\newcommand{\AState}[1]{\State{\smash{#1}}}
\newcommand{\AFor}[1]{\For{\smash{#1}}}

\usepackage{tikz}

\usepackage{amsmath,amsfonts,bm,bbm}

\def\eqref#1{equation~\ref{#1}}
\def\1{\bm{1}}

\def\eps{{\epsilon}}

\def\rvepsilon{{\mathbf{\epsilon}}}

\def\vc{{\bm{c}}}
\def\vd{{\bm{d}}}

\def\vg{{\bm{g}}}

\def\vx{{\bm{x}}}
\def\vy{{\bm{y}}}

\def\mD{{\bm{D}}}

\def\mI{{\bm{I}}}

\def\mR{{\bm{R}}}

\DeclareMathAlphabet{\mathsfit}{\encodingdefault}{\sfdefault}{m}{sl}
\SetMathAlphabet{\mathsfit}{bold}{\encodingdefault}{\sfdefault}{bx}{n}

\def\gN{{\mathcal{N}}}

\def\gR{{\mathcal{R}}}

\def\gX{{\mathcal{X}}}

\newcommand{\E}{\mathbb{E}}

\newcommand\norm[1]{\left\lVert#1\right\rVert}

\newcommand\smallO{
  \mathchoice
    {{\scriptstyle\mathcal{O}}}% \displaystyle
    {{\scriptstyle\mathcal{O}}}% \textstyle
    {{\scriptscriptstyle\mathcal{O}}}% \scriptstyle
    {\scalebox{.7}{$\scriptscriptstyle\mathcal{O}$}}%\scriptscriptstyle
  }

\newcommand{\diff}{\mathrm{d}}

\usepackage{soul,xcolor}

\usepackage{ulem}
\newcommand\redsout{\bgroup\markoverwith{\textcolor{red}{\rule[0.5ex]{2pt}{0.4pt}}}\ULon}

\newcommand{\our}{\texttt{RES}\xspace}
\newcommand{\ourab}{\texttt{RES-AB}\xspace}

\newcommand{\boldzero}{\mathbf{0}}
\newcommand{\boldi}{\mathbf{I}}
\newcommand{\mathcomma}{\scriptstyle{,}}
\newcommand{\hsqrt}[1]{\raisebox{-.0ex}{$\sqrt{\raisebox{0pt}[1.8ex][0.3ex]{$#1$}}$}}

\newcommand{\gd}{g_\mD}
\newcommand{\geps}{g_\eps}
\begin{document}
\title{Improved Order Analysis and Design of Exponential Integrator for Diffusion Models Sampling}
% \author{Qinsheng Zhang
%     \thanks{This work was supported by xxx.
%     }% <-this % stops a space
%     \thanks{Q. Zhang is with Georgia Institute of Technology, Atlanta, GA, USA. {\tt\small qzhang419@gatech.edu}.}
% }
\author{%
  Qinsheng Zhang \\
  Georgia Institute of Technology\\
  \texttt{qzhang419@gatech.edu} \\
\And
  Jiaming Song \\
  Luma AI \\
  \texttt{jiaming.tsong@gmail.com} \\
\And
  Yongxin Chen \\
  Georgia Institute of Technology\\
  \texttt{yongchen@gatech.edu} \\
}

% \author{
% Qinsheng Zhang \And Jiaming Song \And Yongxin Chen
% \And 
% \and{{\tt\small \{qzhang419,yongchen\}@gatech.edu}}\\ 
% Georgia Institute of Technology
% \and{{\tt\small jiaming.tsong@gmail.com}}\\
% Luma AI
% }

\maketitle
\begin{abstract}
    Efficient differential equation solvers have significantly reduced the sampling time of diffusion models (DMs) while retaining high sampling quality. Among these solvers, exponential integrators (EI) have gained prominence by demonstrating state-of-the-art performance. 
    However, existing high-order EI-based sampling algorithms rely on degenerate EI solvers, resulting in inferior error bounds and reduced accuracy in contrast to the theoretically anticipated results under optimal settings.
    This situation makes the sampling quality extremely vulnerable to seemingly innocuous design choices such as timestep schedules. For example, an inefficient timestep scheduler might necessitate twice the number of steps to achieve a quality comparable to that obtained through carefully optimized timesteps.
    To address this issue, we reevaluate the design of high-order differential solvers for DMs. 
    Through a thorough order analysis, we reveal that the degeneration of existing high-order EI solvers can be attributed to the absence of essential order conditions.
    By reformulating the differential equations in DMs and capitalizing on the theory of exponential integrators, we propose refined EI solvers that fulfill all the order conditions, which we designate as Refined Exponential Solver (\our).
    Utilizing these improved solvers, \our exhibits more favorable error bounds theoretically and achieves superior sampling efficiency and stability in practical applications. 
    For instance, a simple switch from the single-step DPM-Solver++ to our order-satisfied \our solver when Number of Function Evaluations (NFE) $=9$, results in a reduction of numerical defects by $25.2\%$ and FID improvement of $25.4\%$ (16.77 vs 12.51) on a pre-trained ImageNet diffusion model.
    % Employing improved EI solvers, \our attains superior sampling efficiency in practice and more favorable error bounds in theory. \yc{also emphasize stability?}
    % \qsh{, for example, it shows up to 40\% acceleration by switching single-step DPM-Solver++ to single-step \our.}\js{maybe find a biggest gap in Figure 2 to support your claims?}
% \yc{For instance, when NFE$=9$, compared with single-step DPM-Solver++, the single step \our reduce $25.2\%$ numerical defects, and $25.4\%$ improvement in FID (16.77 vs 12.51) on pre-trained ImageNet diffusion model.}
% \qsh{Above is for $x0$model, for $\epsilon$ model used in experiment 1. For instance, when NFE$=9$, compared with single-step DPM-Solver++, the single step \our reduce $52.75\%$ numerical defects, and $48.3\%$ improvement in FID (16.77 vs 8.66) on pre-trained ImageNet diffusion model.}
\end{abstract}
% \draftdisclaimer

\section{Introduction}\label{sec:intro}

Diffusion models (DMs)~\cite{ho2020denoising,song2020score} have recently garnered significant interest as powerful and expressive generative models. 
They have demonstrated unprecedented success in text-to-image synthesis~\cite{rombach2022high,saharia2022photorealistic,balaji2022ediffi} and extended their impact to other data modalities, such as 3D objects~\citep{poole2022dreamfusion,lin2022magic3d,shue20223d,bautista2022gaudi}, audio~\citep{kong2020diffwave}, time series~\citep{tashiro2021csdi,bilovs2022modeling}, and molecules~\citep{wu2022protein,qiao2022dynamic,xu2022geodiff}. 
% Compared to other generative models, DMs offer stable training, excellent scalability to large paired data, and flexibility to adapt to various tasks with minimal fine-tuning or even no retraining.
However, DMs suffer from slow generation due to their iterative noise removal process. Generative adversarial networks (GANs) require only a single network evaluation to generate a batch of images, whereas DMs may necessitate hundreds or thousands of network function evaluations (NFEs) to transform Gaussian noise into clean data.

Recently, a surge of research interest has been directed toward accelerating the sampling process in diffusion models.
One strategy involves the distillation of deterministic generation. Despite requiring additional computational resources, this approach can reduce the number of NFEs to fewer than five~\cite{salimans2022progressive,song2023consistency,meng2022distillation}.
Nevertheless, distillation methods often depend on training-free sampling methods during the learning process and are only applicable to the specific model being distilled, thereby limiting their flexibility compared to training-free sampling methods~\cite{luhman2021knowledge}.
Another line of investigation is aimed at designing generic sampling techniques that can be readily applied to any pre-trained DMs. 
Specifically, these techniques leverage the connection between diffusion models and stochastic differential equations (SDEs), as well as the feasibility of deterministically drawing samples by solving the equivalent Ordinary Differential Equations (ODEs) for marginal probabilities, known as probability flow ODEs~\cite{song2020score,song2020denoising}. 
Research in this domain focuses on the design of efficient numerical solvers to expedite the process~\cite{liu2022pseudo,zhang2022fast,lu2022dpm,zhang2022gddim,karras2022elucidating,lu2022dpmp,zhao2023unipc}. 
Despite the significant empirical acceleration, numerous methods that claim the same order of numerical error convergence rate demonstrate significant disparities in practical application. 
Furthermore, these approaches often incorporate a range of techniques and designs that may obscure the actual factors contributing to the acceleration, such as various thresholding~\cite{dhariwal2021diffusion,saharia2022photorealistic}, and sampling time optimization~\cite{deepfloyd23,karras2022elucidating}.~(See \cref{app:related} for more related works and discussions.)

In this study, we initiate by revisiting different diffusion probability flow ODE parameterizations, underlining the vital role of approximation accuracy for integrals in the ODE solution in various sampling algorithms. This directly impacts both the quality and speed of sampling~(\cref{ssec:revist}).
Through a change-of-variable process, we further unify the ODEs for the noise prediction model and data prediction into a singular, canonical semilinear ODE. 
We then conduct an exhaustive order analysis for the single-step numerical scheme for the semilinear ODE and contend that the widely accepted sampling algorithm, which is based on the exponential integrator (EI), contravenes order conditions. This leads to inferior error bounds, compromised accuracy, and a lack of robustness compared to the theoretically projected outcomes under optimal settings~(\cref{ssec:ode-error-analysis}).

In response, we propose Refined Exponential Solver (\our), which offers more favorable error recursion both theoretically and empirically, thus ensuring superior sampling efficiency and robustness~(\cref{ssec:order-condition}). 
Moreover, our single-step scheme can be seamlessly integrated to enhance multistep deterministic~(\cref{ssec:stochastic}) and stochastic sampling algorithms~(\cref{sec:multistep}), surpassing the performance of contemporary alternatives.
% Furthermore, its single-step scheme can be seamlessly incorporated to accelerate stochastic sampling~(\cref{ssec:stochastic}). 
% Drawing inspiration from the EI multistep solver, we introduce the \our multistep sampler~(\cref{sec:multistep}), which demonstrates superior sampling quality in comparison to existing multistep samplers. 
Lastly, we undertake comprehensive experiments with various diffusion models to demonstrate the generalizability, superior efficiency, and enhanced stability of our approach, in comparison with existing works~(\cref{sec:exp}). 
In summary, our contributions are as follows:

\begin{itemize}
\item We revisit and reevaluate diffusion probability flow ODE parameterizations and reveal the source of approximation error. 
\item To minimize error, we propose a unified canonical semilinear ODE, identify the overlooked order condition in existing studies, and rectify this oversight. 
% This lays the groundwork for the development of \our.
% \item We introduce \our, an innovative method that provides improved error recursion, both theoretically and empirically. This results in enhanced sampling efficiency and stability. \qsh{TODO: combine 3/4, add numbers}
% \item Our method is validated through extensive experiments with various diffusion models, underlining its generalizability, superior efficiency, and enhanced stability, as compared to existing methodologies.
\item We introduce \our, an innovative method that provides improved error recursion, both theoretically and empirically. Extensive experiments are conducted to show its generalizability, superior efficiency, and enhanced stability. For example, it shows up to 40\% acceleration by switching single-step DPM-Solver++ to single-step \our under a suboptimal time scheduling.
\end{itemize}
\section{Background}

Given a data distribution of interest, denoted as $p_{data}(\vx)$ where $\vx \in \gX$, a diffusion model consists of a forward noising process that diffuses data point $\vx$ into random noise and a backward denoising process that synthesize data via removing noise iteratively starting from random noise.
We start our discussion with simple variance-exploding diffusion models~\cite{song2020improved}, which can be generalized to other diffusion models under the unifying framework introduced by~\citet{karras2022elucidating}.
Concretely, the forward noising process defines a family of distributions $p(\vx; \sigma(t))$ dependent on time $t$, which is obtained by adding \textit{i.i.d.} Gaussian noise of standard deviation $\sigma(t)$ to noise-free data samples. 
We choose $\sigma(t)$ to be monotonically increasing with respect to time~$t$, such as $\sigma(t) = t$. 

To draw samples from diffusion models, a backward synthesis process is required to solve the following stochastic differential equation~(SDE)~\cite{zhang2021diffusion,huang2021variational,karras2022elucidating}, starting from $\vx(T) \sim \gN(\mathbf{0}, \sigma(T)^2 \mI)$ for a large enough $T$:
\begin{align}
    \diff \vx =& - \underbrace{\dot{\sigma}(t) \sigma(t) \nabla_\vx \log p(\vx; \sigma(t)) \diff  t}_{\text{Probabilistic ODE}} 
    - \underbrace{\beta_t \sigma(t)^2 \nabla_\vx \log p(\vx; \sigma(t)) \diff t + \sqrt{2 \beta_t} \sigma(t) \diff  \omega_t}_{\text{Langevin process}}, \label{eq:score-sde}
\end{align}
where $\nabla_\vx \log p(\vx; \sigma(t))$ is the score function  (\textit{i.e.}, gradient of log-probability), $\omega_t$ is the standard Wiener process, and $\beta_t$ is a hyperparameter that controls the stochasticity of the process. 
\cref{eq:score-sde} reduces to deterministic probabilistic flow ODE~\cite{song2020score} when $\beta=0$.
Some popular methods, such as variance-preserving diffusion models, introduce an additional scale schedule $s(t)$ and consider $\vx = s(t) \hat{\vx}$ to be a scaled version of the original, non-scaled data $\hat{\vx}$. 
Though the introduction of scale schedule $s(t)$ will result in a different backward process, we can undo the scaling and reduce their sampling SDEs to ~\cref{eq:score-sde}. Note that we focus on non-scaling diffusion models in this paper for simplicity and leave the extension to other diffusion models in~\cref{app:proof}.
The score function $\nabla_{\vx} \log p(\vx; \sigma(t))$ of noised data distribution at a particular noise level $\sigma$ can be learned via a denoising score matching objective~\citep{vincent2011a}:
\begin{align}
L_\sigma(\theta) = \E_{\vx \sim p_{data}, \rvepsilon \sim \gN(0, \mI)}[\norm{D_\theta(\vx + \sigma \rvepsilon , \sigma) - \vx}_2^2], \label{eq:dsm}
\end{align}
where $D_\theta: \gX \times \gR \to \gX$ is a time-conditioned neural network that tries to denoise the noisy sample. 
In the ideal case, having the perfect denoiser $D_\theta$ is equivalent to having the score function thanks to Tweedie's formula~\citep{efron2011tweedie}:
\begin{align}
    D_\theta(\vx, \sigma) %\\
     = \vx + \sigma^2 \nabla_{\vx} \log p(\vx; \sigma). \label{eq:x0-pred}
\end{align}
For convenience, we also introduce the noise prediction model, which tries to predict $\epsilon$ in \cref{eq:dsm}:
\begin{align}
    \epsilon_\theta(\vx, \sigma) := \frac{\vx - D_\theta(\vx, \sigma)}{\sigma}. \label{eq:eps-pred}
\end{align}

% \begin{equation}\label{eq:ode}
%     d \vx = \frac{\vx - D_\theta(\vx_t, t)}{\sigma_t} d\sigma_t.
% \end{equation}

\section{Refined exponential solver \our}
% \qsh{some words for a smooth transition from background section to main section, maybe explain why we start with ODE case?}
% \yc{add a few sentences to explain the goal and flow of this section}
Even with the same trained diffusion model, different solvers of \cref{eq:score-sde} will lead to samplers of drastically different efficiency and quality. In this section, we present a unified formulation to the various probability flow ODE being considered (\cref{ssec:revist}), analyze the numerical approximation errors for general single-step methods~(\cref{ssec:ode-error-analysis}), and find the optimal set of coefficients that satisfies the desired order conditions and minimizes numerical error~(\cref{ssec:order-condition}). 
We then extend the improvement to the stochastic setting~(\cref{ssec:stochastic}) and multistep sampling methods~(\cref{sec:multistep}).
% \js{wonder if we need to keep 3.4 and 3.5}
\subsection{Better parameterizations for probability flow ODE}\label{ssec:revist}

The probability flow ODE in diffusion models in \cref{eq:score-sde} is first order, and thus we can draw some analogies between $\vx$ and ``location'', $t$ and ``time'', and $- \dot{\sigma}(t) \sigma(t) \nabla_\vx \log p(\vx, \sigma(t))$ as ``velocity''. In general, an ODE is easier to solve if the ``velocity'' term has smaller first-order derivatives~\cite{press2007numerical,lipman2022flow}. 
Here, we list several different parametrizations for these quantities. While the exact solutions to these ODEs are the same, the numerical solutions can differ dramatically.

\begin{table}[]
\centering
\caption{Different parametrizations of the probabilistic ODE.}
\begin{tabular}{@{}ccccc@{}}
\toprule
Parametrization & Location & Velocity & Time & Semi-linear \\ \midrule
    EDM / DEIS           &  $\vx$        &    $(\vx - \mD_\theta(\vx, t)) / t$      &  $t := \sigma$    &   No          \\
      logSNR          &   $\vx$       &     $- \vx + \mD_\theta(\vx, e^{-\lambda_\mD})$     &   $\lambda_\mD := - \log \sigma$   &   Yes          \\ 
      Negative logSNR          &   $\vy := \vx / e^{\lambda_\eps}$       &        $- \vy + \eps_\theta(e^{\lambda_\eps} \vy, e^{\lambda_\eps})$                     &     $\lambda_\eps := \log \sigma$         & Yes \\\bottomrule
\end{tabular}
\label{tab:param}
\end{table}

\paragraph{EDM} In EDM~\cite{karras2022elucidating}, the location term is on the data space $\vx$, the probability flow ODE in \cref{eq:score-sde} with $\beta_t = 0$ simplifies to the following equation:
\begin{equation}~\label{eq:simple-ode}
    \diff \vx = \frac{\vx - \mD_\theta(\vx, \sigma(t))}{\sigma(t)} \dot{\sigma}(t) \diff t.
\end{equation}
To solve \cref{eq:simple-ode}, ~\citet{karras2022elucidating} interpret the term $\frac{\vx - \mD_\theta(\vx, \sigma(t))}{\sigma_t} \dot{\sigma}(t)$ as a black box function and choose $\sigma(t) = t$, and then apply a standard second-order Heun ODE solver.

\paragraph{DEIS} Similarly, DEIS~\cite{zhang2022fast} uses the noise prediction model $\epsilon_\theta$ to parametrize the ODE:
\begin{equation}~\label{eq:eps-ode}
    \diff \vx = \epsilon_\theta(\vx, \sigma) \diff \sigma.
\end{equation}
It is not hard to see that this is equivalent to the EDM one if we set $\sigma$ to $t$.

\paragraph{logSNR} We explore an alternative probability flow ODE parameterization based on the denoising model $\mD_\theta$. We consider $\lambda_\mD(t) :=- \log \sigma(t)$, which can be treated as the log signal-to-noise ratio (logSNR,~\cite{kingma2021variational}) between coefficients of the signal $\vx(0)$ and noise $\epsilon$. 
With $\diff \lambda_\mD(t) = \frac{-\dot{\sigma}(t)}{\sigma(t)} \diff t$, \cref{eq:simple-ode} can be reformulated as follows:
\begin{equation}\label{eq:semi_ode}
    \frac{\diff \vx}{\diff \lambda_\mD} = - \vx + \gd(\vx, \lambda_\mD), \quad \gd(\vx, \lambda_\mD) := D_\theta(\vx, e^{-\lambda_\mD}).
\end{equation}

\paragraph{Negative logSNR} Alternatively, we can restructure~\cref{eq:eps-ode} by employing a change-of-variables approach with $\vy(t) := \frac{\vx(t)}{\sigma(t)}, \lambda_\eps (t) := \log \sigma(t)$ for the noise prediction model $\epsilon_\theta$:
\begin{equation}\label{eq:semi_ode_eps}
    \frac{\diff \vy}{\diff \lambda_\eps} = - \vy + \geps(\vy, \lambda_\eps),
    \quad \geps(\vy, \lambda_\eps) := \eps_\theta(e^{\lambda_\eps} \vy, e^\lambda_\eps),
\end{equation}
where $\lambda_\eps$ is the negative logSNR.
We observe that ~\cref{eq:semi_ode,eq:semi_ode_eps} possess highly similar ODE structures. In fact, they are both \textit{semilinear parabolic problems} in the numerical analysis literature~\cite{hochbruck2005exponential}, \textit{i.e.}, the velocity term is a linear function of location plus a non-linear function of location ($\gd$ or $\geps$).
This suggests that ODEs with either epsilon prediction or data prediction models can be addressed within a \textit{unified} framework. We summarize these different parametrizations in \cref{tab:param}.

\begin{figure}
    \begin{subfigure}[b]{0.30\textwidth}
        \includegraphics[width=\textwidth]{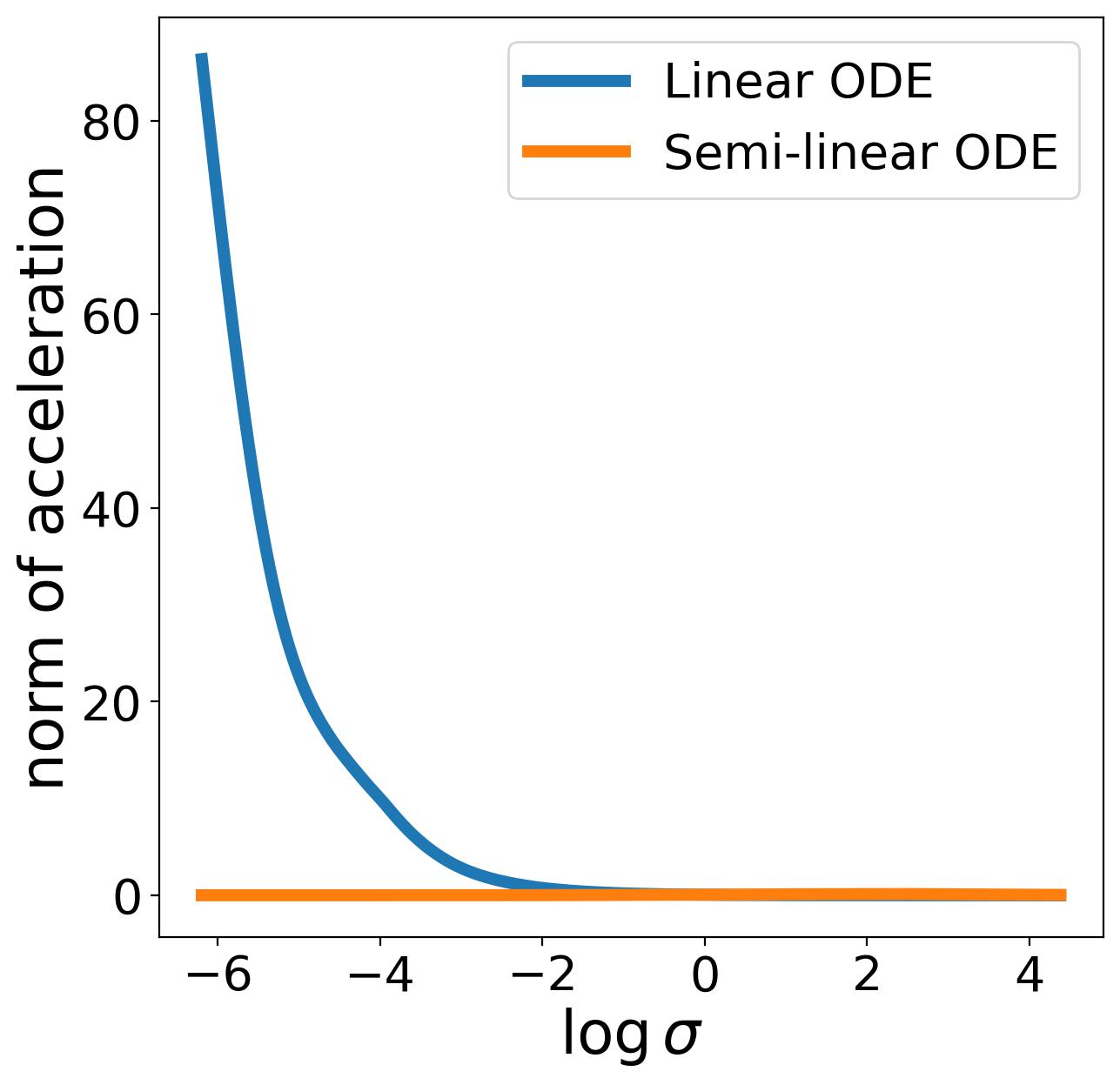}
        \caption{Acceleration vs ODE}
        \label{fig:function-derivative}
    \end{subfigure}
    \hfill
    \begin{subfigure}[b]{0.70\textwidth}
        \includegraphics[width=\textwidth]{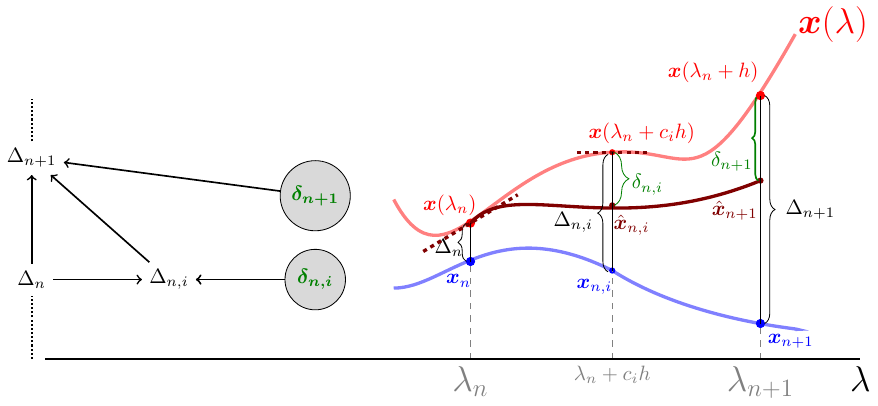}
        \caption{Diagram of error recursion.}
        \label{fig:defect}
    \end{subfigure}
    \caption{
        \cref{fig:function-derivative}. {Along the trajectory of an exact solution to the probability flow ODE, semi-linear ODE with logarithm transformation parametrization on noise level $\sigma$ has smaller ``acceleration'', \textit{i.e.,} the time-derivative of ``velocity''. The curves are averaged over $512$ solutions on the pre-trained ImageNet model~\citep{karras2022elucidating}. This suggests that semi-linear ODE incur less discretization error when solving ODEs.}
        \cref{fig:defect}. 
        The auxiliary trajectory \textcolor{black!50!red}{$\hat{\vx}$} starts from exact solution \textcolor{red} {$\vx$} and is built upon numerical integration proposed in~\cref{eq:sin-scheme}. 
        The diagram depicts that numerical defects $\Delta_{n+1}$ between $\textcolor{red}{{\vx}(\lambda_{n+1})}$ and numerical solution $\textcolor{blue} {\vx_{n+1}}$ is a result of accumulated intermediate defects $\textcolor{black!50!green}{\{\delta_n, \delta_{n,i}\}}$ in each step, which is the discrepancy between the auxiliary trajectory and the exact solution trajectory.
    }
    \label{fig:teaser}
\end{figure}

A notable advantage of the logSNR and negative SNR parametrizations is that their velocity terms along exact solution trajectory, such as 
% \qsh{miss gt solution? $\vx(\lambda_\mD)$ instead of $\vx$}\js{I feel that we might need to include $\vx$ as an argument to $f$? this also makes me wonder how you got figure 1a, is it for one x or an expectation over a distribution of x?})\qsh{just for one $\vx$, then take an average over many $\vx$}\js{migth help to clarify in figure.} 
% \yc{this is wrong $f_\mD(\lambda_\mD) := - \vx(\lambda_\mD) + \gd(\vx(\lambda_\mD),\lambda_\mD)$}
$- \vx(\lambda_\mD) + \gd(\vx(\lambda_\mD),\lambda_\mD)$
, are smoother than the velocity term in the ODE for EDM/DEIS. We illustrate this in~\cref{fig:function-derivative}, which shows that the norm of the derivative of the EDM velocity term grows rapidly as $\sigma \to 0$; 
% \yc{add a reference to the DEIS paper where we have a similar argument} 
this may explain why the timestep scheduling used in popular samplers (such as DDIM and EDM) often place more steps at lower noise levels.

% \begin{figure}[!htp]
%     \centering
%     \includegraphics[width=4cm]{figure/plt/plt/f_derivative_sigma.jpg}
%     \includegraphics[width=4cm]{figure/plt/plt/f_derivative_lambda.jpg}
%     \caption{With logarithm transformation to the noise level $\sigma$, the trajectories of nonlinear function evaluations exhibit improved smoothness. \yc{If only change-of-variable $\lambda = \log \sigma$ is used, then these two plots are not consistent}}
%     \label{fig:function-derivative}
% \end{figure}

% \begin{figure}
%     \begin{subfigure}[b]{0.20\textwidth}
%         \includegraphics[width=\textwidth]{figure/plt/plt/f_derivative_sigma.jpg}
%         \includegraphics[width=\textwidth]{figure/plt/plt/f_derivative_lambda.jpg}
%         \caption{Derivative of $\vf$}
%         \label{fig:function-derivative}
%     \end{subfigure}
%     \hfill
%     \begin{subfigure}[b]{0.78\textwidth}
%         \includegraphics[width=\textwidth]{figure/tikz_pdf/defects.pdf}
%         % \includegraphics[height=5cm]{figure/tikz_pdf/defects.pdf}
%         \caption{Diagram of error recursion.}
%         \label{fig:defect}
%     \end{subfigure}
%     \caption{
%         \cref{fig:function-derivative}. With logarithm transformation to the noise level $\sigma$, the trajectories of nonlinear function evaluations exhibit improved smoothness.
%         \cref{fig:defect}. 
%         The auxiliary trajectory \textcolor{black!50!red}{$\hat{\vx}$} starts from numerical solution \textcolor{blue} {$\vx$} and built upon function evaluations based on exact solution trajectory \textcolor{red}{${\vx}(\lambda)$}. 
%         The diagram depicts that numerical defects $\Delta_{n+1} := \textcolor{red}{{\vx}(\lambda_{n+1})} - \textcolor{blue} {\vx_{n+1}}$ is a result of accumulated intermedia defects $\{\delta_n, \delta_{n,i}\}$ in each step, which is the discrepancy between auxiliary trajectory and exact solution trajectory.
%     }
%     % \caption{Images A, B, and C.}
%     \label{fig:}
% \end{figure}

For the rest of the section, we focus on logSNR~(\cref{eq:semi_ode}); theoretical conclusions and practical algorithms can be easily transferred to negative logSNR~(\cref{eq:semi_ode_eps}).
For notational simplicity, we substitute $\lambda$ for $\lambda_\mD$ and $g$ for $\gd$, so our ODE in \cref{eq:semi_ode} becomes:
\begin{align}
    % \frac{\diff \vx}{\diff \lambda} = \underbrace{- \vx + g(\vx, \lambda)}_{ f(\lambda)},
    \frac{\diff \vx}{\diff \lambda} = - \vx + g(\vx, \lambda),
\end{align}
% \qsh{Sorry, just find definition of $f$ here conflict with the later $f$. Maybe just remove $f$ here.}
% where we use $f(\lambda)$ as a shorthand notation for velocity of exact solution at location $x(\lambda)$ and time $\lambda$.
Suppose that our ODE solver advances for a step size of $h$ from $\vx(\lambda)$, then the exact solution $\vx(\lambda + h)$ to this ODE, as given by the \textit{variation-of-constants} formula, is represented as follows:
\begin{equation}\label{eq:voc}
    \vx(\lambda + h) = e^{-h} \vx(\lambda) + \int_0^{h} e^{(\tau - h)} 
    g(\vx(\lambda+\tau), \lambda+\tau) \diff \tau.
\end{equation}
However, the computation of this exact solution confronts two significant challenges: the intractability of the integral and the inaccessible integrand, which involves the evaluation of the denoiser on the exact solution $\vx(\lambda + \tau)$.
To approximate the intractable integration in~\cref{eq:voc}, we start with the single-step approach and derive its order conditions.

\subsection{Single-step numerical schemes and defect analysis}\label{ssec:ode-error-analysis}
To approximate the exact solution in~\cref{eq:voc}, the numerical scheme of explicit single-step methods, characterized by $s$-stages, can be described as follows: 
% \yc{shouldn't it be $\vd_{n,i} := g(\vx_{n,i}, \lambda_n + c_i h_n)$?}
\begin{subequations}~\label{eq:sin-scheme}
\begin{align}
        \vx_{n+1} &= e^{-h_n} \vx_n + h_n \sum_{i=1}^s b_i(-h_n) \vd_{n,i}, \label{eq:sin-scheme1}\\ 
        \vx_{n,i}  &= e^{-c_i h_n} \vx_n + h_n \sum_{j=1}^{i-1} a_{ij}(-h_n) \vd_{n,j}, \quad
        \vd_{n,i} := g(\vx_{n,i}, \lambda_n + c_i h_n),\label{eq:sin-scheme2}
\end{align}
\end{subequations}
where $h_n:=\lambda_{n+1} - \lambda_n$ denotes the step size from state $\vx_n$ to $\vx_{n+1}$, $\vx_{n,i}$ denotes the numerical solution at the $i$-th stage with timestep $\lambda_n + c_i h_n$, and $\vd_{n,i}$ is the function evaluation at the $i$-th stage.
Notably, $\vx_n$ denotes the numerical solution at time $\lambda_n$, which deviates from the exact solution $\vx(\lambda_n)$; similarly, the numerical solution $\vx_{n,i}$ at stage $i$ deviates from the exact $\vx(\lambda_n + c_i h_n)$. 
The coefficients ${a_{ij},b_i,c_j}$ in~\cref{eq:sin-scheme} can be compactly represented using Butcher tableaus \cref{tab:butcher_tableau_example}~\cite{press2007numerical}.
% For explicit methods, we have $c_1=0$.
% \yc{provide an example of ${a_{ij},b_i,c_j}$ for 1st order solver}
\begin{table}[!h]
\centering
\caption{
        \textit{Left:} Tableau form of $\{a_{ij},b_i,c_j\}$ for numerical scheme in~\cref{eq:sin-scheme}. For all $1 \leq j \leq i \leq s$, $a_{ij} = 0$. \textit{Middle:} For Euler method, $c_1 = 0, b_1 = 1$. \textit{Right:} For Heun's method, $c_1 = 0, c_2 = 1, a_{21}=1, b_1 = b_2 = 0.5$. Unlike our \our exponential integrator, the explicit Euler and Heun solvers do not have the additional exponential coefficients (\textit{e.g.}, $e^{-h_n}$) in front of $\vx_n$ (see \cref{eq:sin-scheme}). These coefficients are not reflected in the Butcher tableau.
        % \yc{how about $b, a$?}.
    }
    \vspace{0.5em}
\begin{subtable}{0.35\textwidth}
    \centering
    $\begin{array}
            {c|cccc}
            c_1\\
            c_2 & a_{21} \\
            \vdots & \vdots & \ddots \\
            c_s & a_{s1} & \cdots & a_{s,s-1} \\
            \hline
                & b_1 & \cdots & b_{s-1} & b_s
    \end{array}$
\end{subtable}
~
\begin{subtable}{0.2\textwidth}
    \centering
    $\begin{array}
            {c|c}
            0 &  \\
            \hline
             & 1
    \end{array}$
\end{subtable}
~
\begin{subtable}{0.3\textwidth}
    \centering
    $\begin{array}
            {c|cc}
            0 & \\
            1 & 1 & \\
            \hline
                & 0.5 & 0.5
    \end{array}$
\end{subtable}
    \label{tab:butcher_tableau_example}
\end{table}

To bound the numerical defect $\vx_n - \vx(\lambda_n)$, we first decompose it into the summation of, $\vx_n - \hat{\vx}_n$ and $\hat{\vx}_n - \vx(\lambda_n)$. Here, the auxiliary trajectory $\hat{\vx}_n$ is obtained by substituding $\vd_{n,i}$ in~\cref{eq:sin-scheme} with 
$f(\lambda) := g(\vx(\lambda), \lambda)$. 
Specifically, we define \textit{intermediate numerical defects} $\delta_{n}, \delta_{n\scriptstyle{,}i}$ as 
% \yc{below is not consistent with (11)}
\begin{subequations}
    \begin{align}
        \scalebox{0.97}{$\delta_{n\scriptstyle{,}i} :=\vx(\lambda_n + c_i h_n) - \hat{\vx}_{n,i}, \quad
        \hat{\vx}_{n\mathcomma i} = e^{-c_i h_n} \vx(\lambda_n) - h_n \sum_{j=1}^{i-1} a_{ij}(-h_n) f(\lambda_n + c_j h_n),$} \label{eq:single-step-intermediate} \\
        \scalebox{0.97}{$\delta_{n+1} :=\vx(\lambda_{n+1}) - \hat{\vx}_{n+1}, \quad \hat{\vx}_{n+1} =
        e^{-h_n} \vx(\lambda_n) - h_n \sum_{i=1}^s b_i(-h_n) f(\lambda_n + c_i h_n).$} \label{eq:single-step-final}
    \end{align}
\end{subequations}
Intuitively, defects $\delta_{n\scriptstyle{,}i}, \delta_{n+1}$ are caused by approximating the intractable integration in~\cref{eq:voc} by finite summation in~\cref{eq:sin-scheme}.
Let $\Delta_n := \vx_n - \vx(\lambda_n)$, $\Delta_{n\mathcomma i} := \vx_{n\mathcomma i} - \vx(\lambda_n + c_i h_n)$, we can derive error recursion with the help of $\delta_{n\scriptstyle{,}i}, \delta_{n+1}$, such that
\begin{subequations}\label{eq:single-error-recursion}
    \begin{align}
    \scalebox{0.95}{$\Delta_{n+1} = e^{-h_n} \Delta_n + h \sum_{i=1}^s b_i(-h_n) (g(\vx_{n\mathcomma i}, \lambda_n + c_i h_n) - f(\lambda_n + c_i h_n)) - \delta_{n+1}$,} \label{eq:error-recursion-final}\\
    \scalebox{0.95}{$\Delta_{n\mathcomma i} = e^{-c_i h_n} \Delta_n + h \sum_{j=1}^{i-1} a_{ij}(-h_n) 
        (g(\vx_{nj}, \lambda_n + c_j h_n) - f(\lambda_n + c_j h_n)) - \delta_{n\scriptstyle{,}i} $.}
    \end{align}
\end{subequations}
\cref{eq:single-error-recursion} reveals that numerical defects $\Delta_{n+1},\Delta_{n\mathcomma i}$ are influenced by the emergent intermediate defects $\delta_{n+1}, \delta_{n\scriptstyle{,}i}$ and defects $\Delta_{n}$, which are themselves modulated by previous intermediate defects. 
Moreover, the discrepancy $g(\vx_{n\mathcomma i}, \lambda_n + c_i h_n) - f(\lambda_n + c_i h_n)$ is contingent upon $\Delta_{n\mathcomma i}$.
As error recursion illustrated in~\cref{fig:defect}, the origin of final numerical defects lies in the defects $\delta_{n+1}, \delta_{n\scriptstyle{,}i}$ in each step. 
Consequently, to curtail the ultimate defect, it becomes paramount to constrain $\delta_{n+1}, \delta_{n\scriptstyle{,}i}$.
To establish bounds for $\Delta_n$ using the above error recursion, the following lemma provides an expansion of intermediate defects, uncovering the association between $\delta_{n+1}, \delta_{n\scriptstyle{,}i}$ and the derivative~of~$f(\lambda)$.
\begin{lemma}\label{lemma:intermediate-defect}
    Let $\phi_{j}(-h_n) = \frac{1}{h_n^j} \int_{0}^{h_n} e^{\tau - h_n} \frac{\tau^{j-1}}{(j -1)!} d \tau$, the order-$q$ Taylor expansion of the intermediate defects $\delta_{n\scriptstyle{,}i}, \delta_{n+1}$ can be formulated as follows:
    \begin{equation}
        \delta_{n\scriptstyle{,}i} = \sum_{j=1}^{q} h_n^j \psi_{j,i}(-h_n) f^{(j-1)}(\lambda_n) + \smallO(h_n^q)
        \quad
        \delta_{n+1} = \sum_{j=1}^{q} h_n^j \psi_{j}(-h_n) f^{(j-1)}(\lambda_n) + \smallO(h_n^q),
    \end{equation}
    where function $\psi_{j,i}, \psi_j$ for the coefficients of the $(j-1)$-th order derivative of $f(\lambda_n)$ are defined as
    \begin{equation*}
        \psi_{j,i}(-h_n) = \phi_j(-c_i h_n ) c_i^j - \sum_{k=1}^{i-1} a_{ik}(-h_n) \frac{c_k^{j-1}}{(j-1)!},
        \quad
        \psi_j(-h_n) = \phi_j(-h_n) - \sum_{k=1}^s b_k(-h_n) \frac{c_k^{j-1}}{(j-1)!}.
    \end{equation*}
\end{lemma}
\Cref{lemma:intermediate-defect} furnishes an expansion expression for $\delta_{n\scriptstyle{,}i},\delta_{n+1}$, with each term contingent upon the derivative of $f$ and coefficients $\{\psi_j, \psi_{j,i} \}$. 
This implies that to minimize numerical defects ($\Delta_{n+1}$), an ODE parameterization that ensures the smoothest possible nonlinear function along the exact solution trajectory should be selected, thus substantiating our choice of the logarithmic transformation on the noise level. To this end, we can constrain numerical defects $\delta_{n+1}, \delta_{n\scriptstyle{,}i}$ by minimizing the magnitude of $\{\psi_j, \psi_{j,i} \}$, forming the foundation of the order conditions for the single-step scheme.

\subsection{Order conditions}\label{ssec:order-condition}
With~\cref{eq:single-error-recursion} and \Cref{lemma:intermediate-defect}, we can deduce the conditions necessary to achieve a numerical scheme of order $q$. We start with the case of a single-stage solver of order one.
\begin{thm}[Error bound for solvers that satisfy the 1st-order condition] \label{thm:order1}
    When $\psi_1(-h_i) = 0$ is satisfied for $1 \leq i \leq n$, the error bound of first-order solver based on~\cref{eq:sin-scheme}
    % \yc{I am confused with the result. If $g'=0$, i.e., $g$ is constant, then the EI should have 0 error. In this case, $f'$ is not 0.}
    \begin{equation}
        \norm{\vx_n - \vx(\lambda_n)} \leq Ch \sup_{\lambda_{\text{min}} \leq \lambda \leq \lambda_\text{max}} \norm{f^\prime(\lambda)}
    \end{equation}
    holds for $h = \max_{1 \leq i \leq n} h_i$. The constant $C$ is independent of $n,h$.
\end{thm}
With~\cref{thm:order1}, $b_1(-h)=\phi_1(-h)$ and the numerical scheme reads
\begin{equation}~\label{eq:ddim}
    \vx_{n+1} = e^{-h_n} \vx_n + h_n \phi_1(-h) g(\vx_n, \lambda_n),
\end{equation}
which is known as exponential Euler~\cite{hochbruck2005exponential} or DDIM for diffusion model~\cite{song2020denoising}.
Although our order analysis aligns with existing work~\cite{lu2022dpm,zhang2022fast}, the distinction emerges in high-order methods.
\begin{thm}[Error bound for solvers that satisfy the 1st- and 2nd-order conditions]\label{thm:order2}
    When $\psi_1(-h_i) =\psi_2(-h_i) = \psi_{1,2}(-h_i) = 0$ is satisfied for $1 \leq i \leq n$, the error bound of second order solver based on~\cref{eq:sin-scheme} 
    \begin{equation}
        \norm{\vx_n - \vx(\lambda_n)} \leq Ch^2 (
        \sup_{\lambda_{\text{min}} \leq \lambda \leq \lambda_\text{max}} \norm{f^\prime(\lambda)} + 
        \sup_{\lambda_{\text{min}} \leq \lambda \leq \lambda_\text{max}} \norm{f^{\prime \prime}(\lambda)})
    \end{equation}
    holds for $h = \max_{1 \leq i \leq n} h_i$. The constant $C$ is independent of $n,h$.
\end{thm}

% To obey~\cref{thm:order2}, coefficients in Butcher tableau have to satisfy
% \begin{equation}
%     b_1(-h \mA) + b_2(-h \mA) = \phi_1(-h \mA),
%     \quad
%     b_2(-h \mA) c_2  = \phi_2(-h \mA),
%     \quad
%     a_{21}(-h \mA) = c_2 \phi_{1,2},
% \end{equation}
% where
% \begin{equation}\label{eq:abbr-phi}
%     \phi_{i,j} = \phi_{i,j}(-h A) = \phi_i(-c_j hA), \quad 2\leq j \leq s.
% \end{equation}
To satisfy order conditions in~\cref{thm:order2}, the optimal Butcher tableau can be parameterized by $c_2$ as shown in~\cref{tab:butcher-tableau}.
It turns out the single-step scheme proposed in DPM-Solver++~\cite{lu2022dpmp} can also be reformulated with a different Butcher tableau as shown in~\cref{tab:butcher-tableau}~(See \cref{app:proof}).
Nevertheless, compared with~\cref{thm:order2}, DPM-Solver++ breaks the order condition $\psi_{2}(-h) =0$. Therefore, DPM-Solver++ does not fully satisfy the full order condition and introduces additional errors in solving ODEs~(See~\cref{app:proof}) theoretically. 
Practically, we find our numerical scheme that obeys the order conditions admits smaller numerical defects and generates higher sample quality from diffusion models. Single-step schemes for orders higher than 3 can also be derived, but we leave them for future work.
% \yc{why future work? we can add it to the appendix} \qsh{The high orders do not improve much according to previous experience, and derivation is more complex/ much longer even for 4.} 
% \yc{then the following should not be a theorem}

\begin{algorithm}[t]
    \footnotesize
    \captionof{algorithm}[single-step]{\atphantom\ \our Second order Single Update Step with $c_2$}
    \begin{spacing}{1.1}
    \begin{algorithmic}[1]
      \AProcedure{SingleUpdateStep}{$\vx_i, \sigma_i, \sigma_{i+1}$}
            \AState{$\lambda_{i+1}, \lambda_i \gets -\log(\sigma_{i+1}),  -\log( \sigma_i)$}
            \AState{$h \gets \lambda_{i+1} - \lambda_i$}
                \AComment{Step length}
            \AState{$a_{21}, b_1, b_2 \gets$ \cref{tab:butcher-tableau} with $c_2$}
                \AComment{Runge Kutta coeffcients}
          % \AState{$\vd_{i,1} \gets D_\theta(\vx_i, \lambda_i) -\vx_i $}
                % \AComment{Evaluate $\diff\vx / \diff \lambda$ at $(\vx,\,\lambda_i)$}
          \AState{$(\vx_{i,2},\,\lambda_{i,2}) \gets (e^{-c_2 h} \vx_i + a_{21} h D_\theta(\vx_i, \lambda_i),\,\lambda_i + c_2 h)$}
                \AComment{Additional evaluation point}
          % \AState{$\vd_{i,2} \gets D_\theta(\vx_{i,2}, \lambda_{i,2}) -\vx_{i,2} $}
                % \AComment{Evaluate $\diff\vx / \diff \lambda$ at $(\vx_{i,2},\,\lambda_{i,2})$}
          \AState{$\vx_{i+1} \gets e^{-h}\vx_i + h(b_1 D_\theta(\vx_i, \lambda_i) + b_2 D_\theta(\vx_{i,2}, \lambda_{i,2}))$}
        \AState{\textbf{return} $\vx_{i+1}$}
      \EndProcedure
    \end{algorithmic}
    \end{spacing}
    \label{alg:eis-single-second}
\end{algorithm}

% \begin{subequations}
%     \begin{align}
%         b_1(-h \mA) + b_2(-h \mA) &= \phi_1(-h \mA) \label{eq:eis-a}\\
%         b_2(-h \mA) c_2  &= \phi_2(-h \mA) \label{eq:eis-b}\\
%         a_{21}(-h \mA) = c_2 \phi_1(-c_2 h \mA) \label{eq:eis-c}
%     \end{align}
% \end{subequations}

% Existing work, which is based on ~\cref{eq:dpmpp-single-2}, break order condition~\cref{eq:eis-b} and resulted in degenerated solution
% \begin{equation}\label{eq:dpmpp-single-2}
%     \begin{array}{c|cc}
%          0 \\
%          c_2 & c_2 \phi_{1,2} \\
%          \hline
%          0 & (1 - \frac{1}{2 c_2}) \phi_1 & \frac{1}{2 c_2} \phi_1
%     \end{array}
% \end{equation}

\begin{table}[htbp]
    \centering
    \caption{Butcher tableau comparison between \our and DPM-Solver++~\cite{lu2022dpmp} for step size $h$, where $\phi_i := \phi_{i}(-h)$ is defined in \Cref{lemma:intermediate-defect}, and we further define $\phi_{i,j} := \phi_{i,j}(-h ) = \phi_i(-c_j h)$. ${\dag}$: $\gamma$ in third order \our needs to satisfy $2(\gamma c_2 + c_3) = 3(\gamma c_2^2 + c_3^3)$. 
    The second order tableau is completely characterized by $c_2$, whereas the third order is parameterized by both $c_2$ and $c_3$.
    % \yc{It is not clear how to choose ${a_{ij},b_i,c_j}$ from this table. What's the relation between $a,b,c$? We cannot assume the reader is an expert on numerical ODE}
    }
    \begin{tabular}{m{1.5cm}|l|l|l}
        \toprule
        & Second Order & Third Order$^{\dag}$ & Order conditions\\
        \midrule
        \our & \scalebox{0.8}{$
                \begin{array}{c|cc}
                    0 \\
                    c_2 & c_2 \phi_{1,2} \\
                    \hline
                    0 & \phi_1 - \frac{1}{c_2} \phi_2 & \frac{1}{c_2} \phi_2
                \end{array}
            $}
            & \scalebox{0.7}{$
            \renewcommand\arraystretch{1.0}
            \begin{array}
            {c|ccc}
            0\\
            c_2 & c_2 \phi_{1,2} \\
            c_3 & c_3 \phi_{1,3} - a_{32} & \gamma c_2 \phi_{2,2} + \frac{c_3^2}{c_2} \phi_{2,3} \\
            \hline
                & \phi_1 - b_2 - b_3 & \frac{\gamma}{\gamma c_2 + c_3} \phi_2 & \frac{1}{\gamma c_2 + c_3} \phi_2
            \end{array}
            $} 
            & {\color{green!50!black} Yes}\\
            \midrule
            DPM Solver++
            & \scalebox{0.8}{$
                \begin{array}{c|cc}
                    0 \\
                    c_2 & c_2 \phi_{1,2} \\
                    \hline
                    0 & (1 - \frac{1}{2 c_2}) \phi_1 & \frac{1}{2 c_2} \phi_1
                \end{array}
            $} 
            & \scalebox{0.7}{$
            \renewcommand\arraystretch{1.0}
            \begin{array}
            {c|ccc}
            0\\
            c_2 & c_2 \phi_{1,2} \\
            c_3 & c_3 \phi_{1,3} - a_{32} & \frac{c_3^2}{c_2} \phi_{2,3} \\
            \hline
                & \phi_1 - b_2 - b_3 & 0 & \frac{1}{c_3} \phi_2
            \end{array}
            $} 
            & {\color{red!50!black} Degenerate}\\
        \bottomrule
    \end{tabular}
    \label{tab:butcher-tableau}
\end{table}

\begin{prop}[(Informal)]~\label{prop:order3}
    Employing the Butcher tableau outlined in~\cref{tab:butcher-tableau}, the single-step method~\cref{eq:sin-scheme}, characterized by three stages, functions as a third-order solver.
    % \begin{equation}\label{eq:deispp-single-3}
    %     \renewcommand\arraystretch{1.0}
    %     \begin{array}
    %     {c|ccc}
    %     0\\
    %     c_2 & c_2 \phi_{1,2} \\
    %     c_3 & c_3 \phi_{1,3} - a_{32} & \eta c_2 \phi_{2,2} + \frac{c_3^2}{c_2} \phi_{2,3} \\
    %     \hline
    %         & \phi_1 - b_2 - b_3 & \frac{\eta}{\eta c_2 + c_3} \phi_2 & \frac{1}{\eta c_2 + c_3} \phi_2
    %     \end{array}
    % \end{equation}
    % where $2(\eta c_2 + c_3) = 3 (\eta c_2^2 + c_3^3)$.
\end{prop}

\subsection{Single-step update for deterministic and stochastic sampler}
\label{ssec:stochastic}

With the single-step update scheme developed in~\cref{ssec:order-condition}, we are prepared to develop a deterministic sampling algorithm by iteratively applying~\cref{eq:sin-scheme} with the corresponding Butcher tableau. Inspired by the stochastic samplers proposed in~\citet{karras2022elucidating}, which alternately execute a denoising backward step and a diffusion forward step, we can replace the original Heun-based single-step update with our improved single-step method. 
Concretely, one update step of our stochastic sampler consists of two substeps. 
The first substep simulates a forward noising scheme with a relatively small step size, while the subsequent substep executes a single-step ODE update with a larger step.
We unify both samplers in~\cref{alg:eis-single} where the hyperparameter $\eta$ determines the degree of stochasticity.
\begin{algorithm}[t]
    \footnotesize
    \captionof{algorithm}[single-step]{\atphantom\ \our Single-Step Sampler}
    \begin{spacing}{1.1}
    \begin{algorithmic}[1]
      \AProcedure{SingleSampler}{$D_\theta, ~\sigma_{i \in \{0, \dots, N\}},~\eta_{i \in \{0, \dots, N\}}$}
        \AState{{\bf sample} $\vx_0 \sim \mathcal{N} \big( \mathbf{0}, ~\sigma^2_0 \mathbf{I} \big)$}
            \AComment{Generate initial sample at $\sigma_0$}
        \AFor{$i \in \{0, \dots, N-1\}$}
        % \AComment{$\gamma_i = \begin{cases}\scalebox{0.9}{${\min}\Big( \frac{\Schurn}{N}, \scalebox{0.8}{$\sqrt{2}{-}1$} \Big)$} & \text{if } \scalebox{0.9}{$t_i{\in}[\Stmin{,}\Stmax]$} \\ 0 & \text{otherwise}\end{cases}$}
        \AState{{\bf sample} $\boldsymbol{\epsilon}_i \sim \mathcal{N} \big( \boldzero, ~\boldi \big)$}
        \AState{$\bar \sigma_i \gets  \sigma_i + \eta_i  \sigma_i$}
            % \AComment{Substep 1.  $\bar t_i$}
        \AState{$\bar{\vx}_i \gets \vx_i + \hsqrt{\bar \sigma_i^2 - \sigma_i^2} ~\boldsymbol{\epsilon}_i$}
            \AComment{Move from $\sigma_i$ to $\bar \sigma_i$ via adding noise}
        \AState{${\vx}_{i+1} \gets $ SingleUpdateStep $(\bar{\vx}_i, \bar{\sigma}_i, \sigma_{i+1})$}
            \AComment{Run single update step from $\bar \sigma_i$ to $\sigma_{i+1}$}
        \EndFor
        \AState{\textbf{return} $\vx_N$}
            \AComment{Return noise-free sample at $t_N$}
      \EndProcedure
    \end{algorithmic}
    \end{spacing}
    \label{alg:eis-single}
\end{algorithm}

\subsection{Multi-step scheme}
\label{sec:multistep}
Instead of constructing a single-step multi-stage numerical scheme to approximate the intractable integration in~\cref{eq:voc}, we can employ a multi-step scheme that capitalizes on the function evaluations from previous steps. Analogous to the analysis in~\cref{ssec:ode-error-analysis} and conclusions in~\cref{thm:order2}, we can extend these results to a multi-step scheme, the details of which are provided in~\cref{app:algo}. 
% \yc{present a bit more details in the main paper if we have space}
Intuitively, multi-step approaches implement an update similar to~\cref{eq:sin-scheme1}. Instead of selecting intermediate states and incurring additional function evaluations, a multi-step method reuses function evaluation output from previous steps.
In this work, our primary focus is on the multi-step predictor case. However, our framework and analysis can also be applied to derive multi-step corrector, which has been empirically proven to further improve sampling quality~\cite{zhao2023unipc,li2023era}.

\section{Experiments}\label{sec:exp}

We further extend our experiments to address the following queries:
(1) Do the deterministic numerical schemes of~\our outperform existing methods in terms of reduced numerical defects and enhanced robustness?
(2) Does a reduction in numerical defects translate to improved sampling quality?
(3) Can our single-step numerical scheme enhance the performance of existing stochastic samplers?
(4) How does ~\our perform across various unguided and guided diffusion models?

\paragraph{Numerical defects for deterministic sampler}

\begin{figure}[h!]
    \centering
   \includegraphics[width=0.27\textwidth, height=0.27\textwidth]{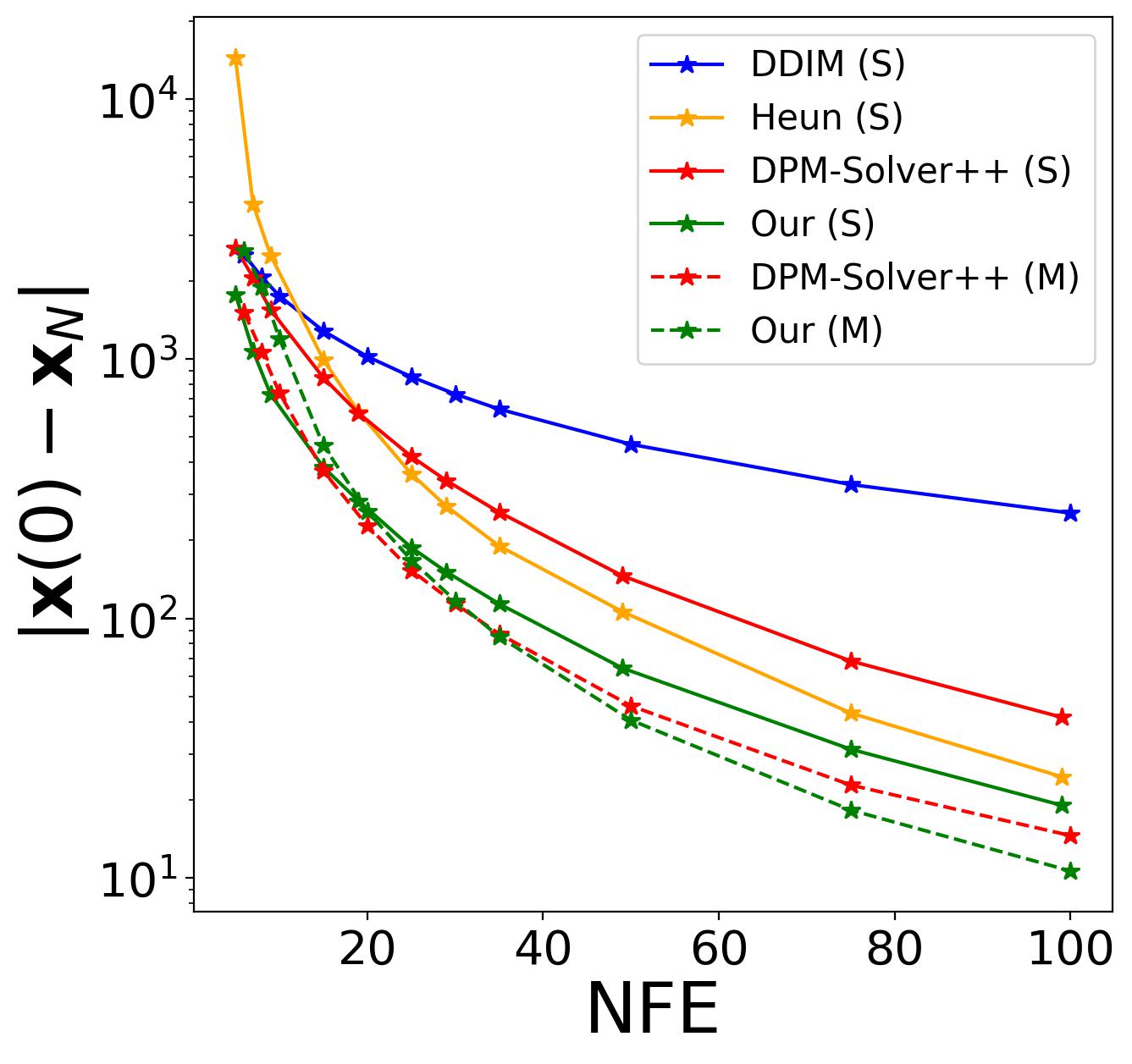}
   \includegraphics[width=0.27\textwidth, height=0.27\textwidth]{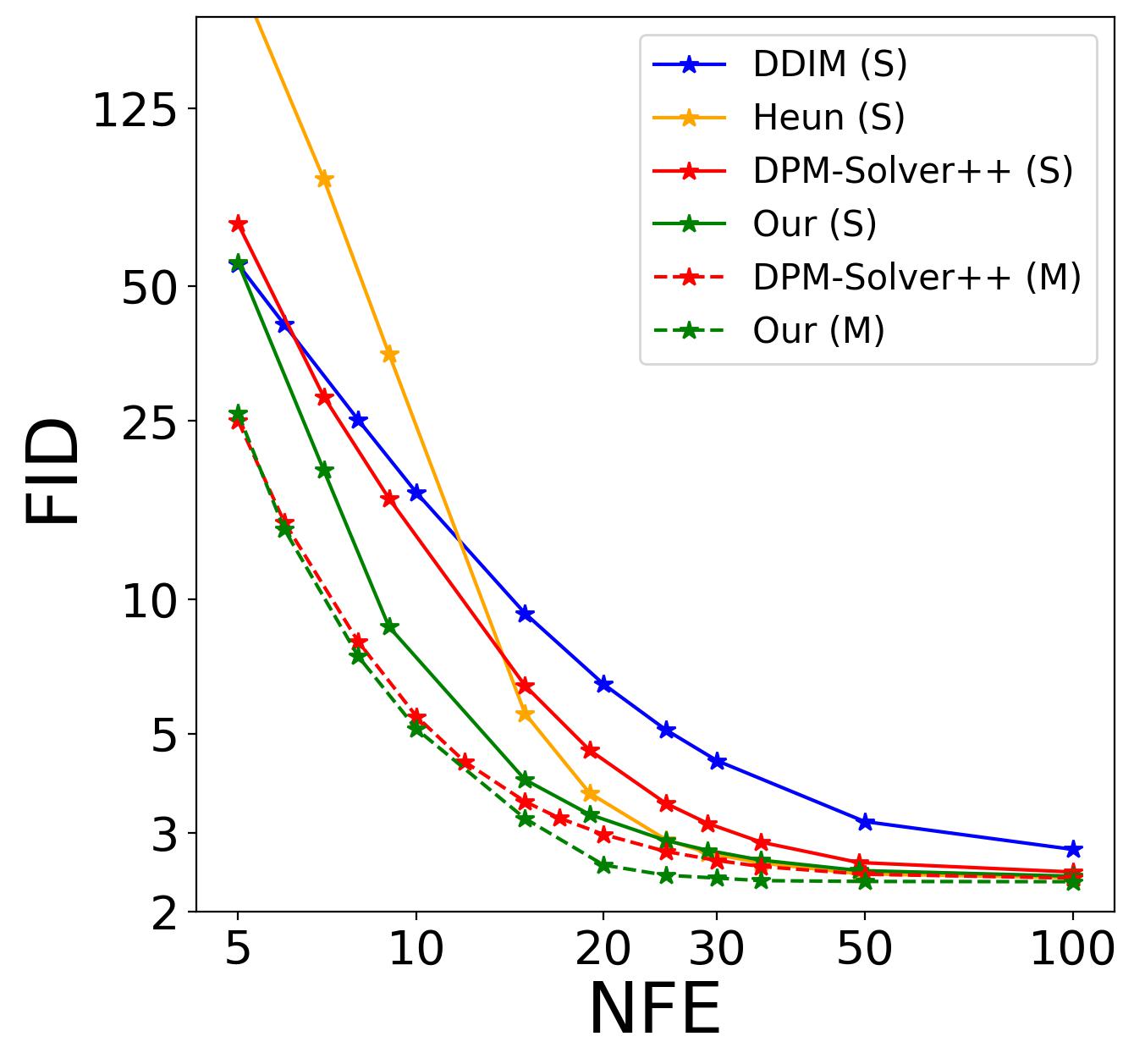}
   \includegraphics[width=0.27\textwidth, height=0.27\textwidth]{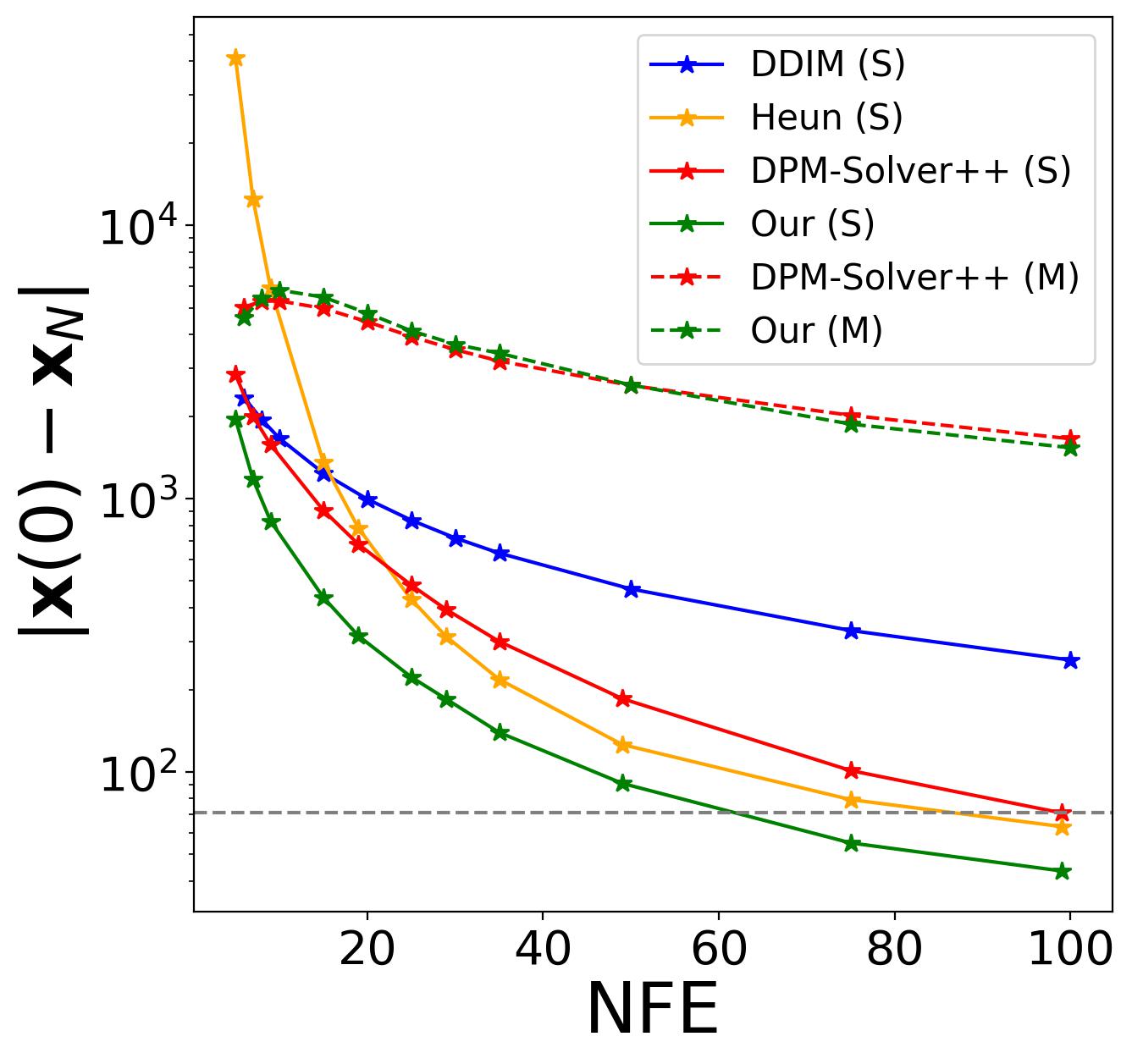}
   \caption{ Comparison among various deterministic samplers on pretrained ImageNet diffusion model~\cite{karras2022elucidating}. $(\text{S})$ indicates single-step methods while $(\text{M})$  for multistep methods.
   (Left) Numerical defects of various sampling algorithms \textit{vs} the number of function evaluation~(NFE) with recommended time scheduling~\cite{karras2022elucidating}. (Middle) FID of various sampling algorithms \textit{vs} NFE with recommended time scheduling. (Right) Numerical defects \textit{vs} NFE with suboptimal time schedule. \our $(\text{S})$ shows better robustness against suboptimal-scheduling. Remarkably, with only 59 NFE, the single-step \our attains a numerical accuracy on par with the 99 NFE DPM-Solver++~(S).}
   % \js{legend of middle image? also maybe use RES (ours) instead of just Our}
   \label{fig:numerical_imagenet}
\end{figure}

First, we investigate the evolution of numerical defects $\norm{\vx_N - \vx(0)}$ in $L$1 norm as the number of function evaluations (NFE) increases in pre-trained ImageNet diffusion models with default hyperparameters. 
Since the exact solution $\vx(0)$ is unavailable, we approximate it using a 500-step RK4 solution, which exhibits negligible changes with additional steps. 
For a fair comparison, we evaluate the \textbf{second-order \our} against single-step DPM-Solver++(S)\cite{lu2022dpmp} and Heun~\cite{karras2022elucidating}, both claimed to be second-order solvers. 
We also include first order DDIM~\cite{song2020denoising} as a baseline.
Our findings indicate that single-step \our exhibits significantly smaller numerical defects, consistent with our theoretical analysis. 
\our based on the noise prediction model (negative logSNR) surpasses the data prediction model (logSNR) in performance for guidance-free diffusion models; we provide a detailed presentation of the former here~(More details in \cref{app:expr}).
% Notably, the \our method based on the $\epsilon$ parameterization outperforms others when the NFE is large, while the $\mD$ parameterization yields smaller errors for relatively small NFE values. 
Additionally, we compare single-step solvers with multistep solvers, specifically including second-order multistep \our and DPM-Solver++~(M). 
We observe \ourab consistently outperforms DPM-Solver++~(M) and both lead to smaller defects compared with single-step methods.
We observe that multistep solvers initially reduce numerical defects quickly but display smaller reductions for larger NFE values. We hypothesize that multistep solvers benefit from twice the number of steps compared to single-step methods when the NFE is relatively small.

\paragraph{Numerical defects and FID} We further investigate the relationship between numerical defects and image sampling quality, using the Frechet Inception Distance (FID)~\cite{heusel2017gans}. As shown in \cref{fig:numerical_imagenet}, smaller numerical defects generally lead to better FID scores as the number of function evaluations (NFE) increases for all algorithms. However, we also observe that the correlation may not be strictly positive and similar defects may lead to different FIDs. We observed a Pearson correlation coefficient of $0.956$ between these two metrics, which suggests that better FID scores are \textit{strongly correlated} with smaller numerical defects.
Notably, when NFE $=9$, compared with single-step DPM-Solver++, the single step \our with noise prediction model (negative logSNR) reduce $52.75\%$ numerical defects, and $48.3\%$ improvement in FID (16.77 vs 8.66). For \our with data prediction model (logSNR), \our achieves $25.2\%$ numerical defects and $25.4\%$ FID improvement (16.77 vs 12.51).

\paragraph{Time-scheduling robustness}
We also evaluate the performance of these solvers concerning varying timestep schedules. We observe that existing methods lack systematic strategies for optimal timestep selection, often relying on heuristic choices or extensive hyperparameter tuning. An ideal algorithm should demonstrate insensitivity to various scheduling approaches. Instead of using the recommended setting in EDM~\cite{karras2022elucidating}, we test these algorithms under a suboptimal choice with uniform step in $\sigma$. While all algorithms exhibit decreased performance under this setting, the single-step \our method outperforms the rest. Notably, single-step methods surpass multi-step methods in this scenario, which signifies the robustness of single-step \our and underscores the benefits of our principled single-step numerical scheme.
Besides, to achieve a numerical accuracy comparable to that of the 99 NFE DPM-Solver++~(S), our single-step method requires only 59 NFE.

\paragraph{Classifier-guided DMs}
\begin{figure}[h!]
    \centering
   \includegraphics[width=0.24\textwidth, height=0.22\textwidth, trim={0 1.5cm 0 0},clip]{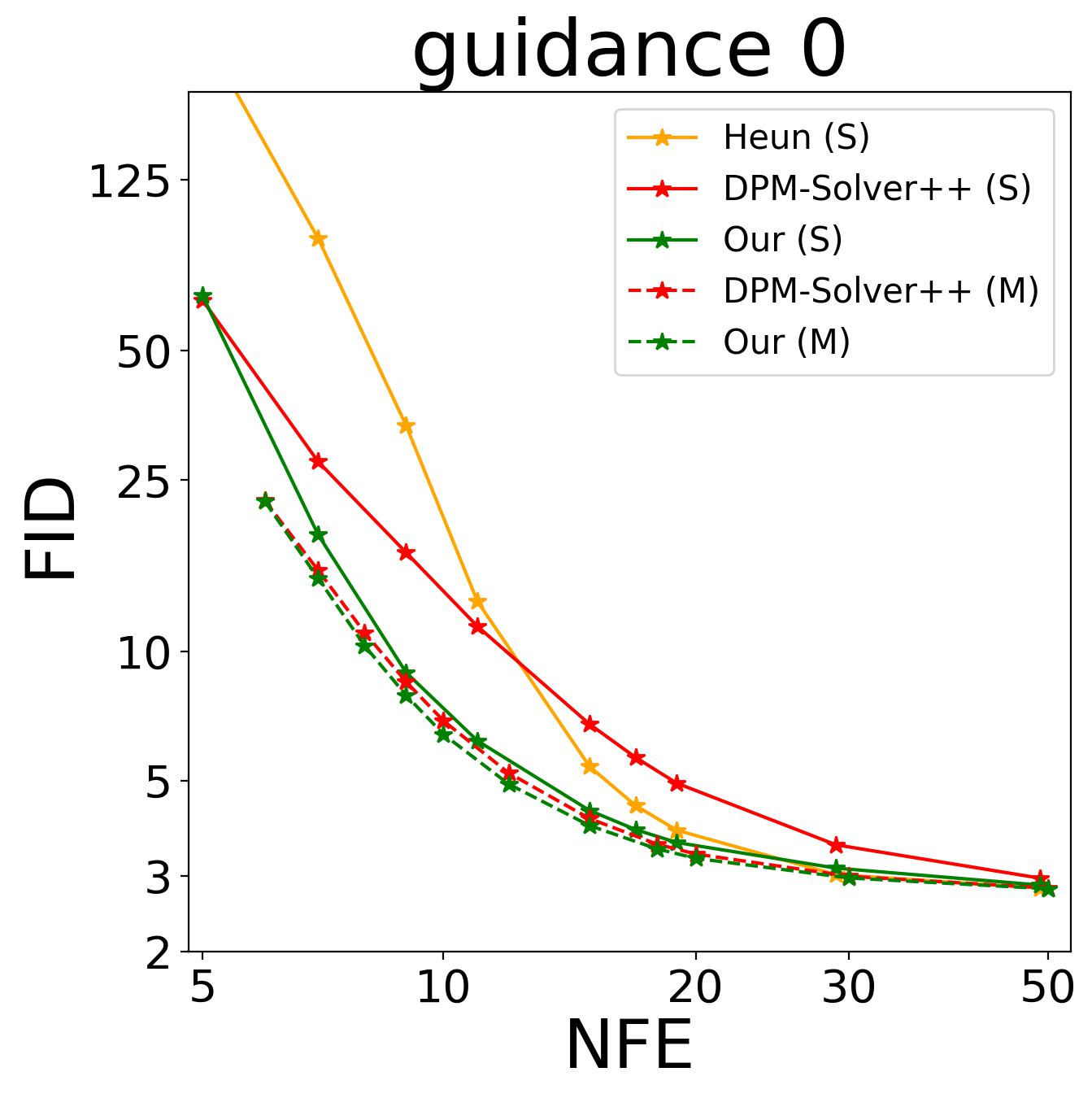}
\includegraphics[width=0.24\textwidth, height=0.22\textwidth, trim={0 1.5cm 0 0},clip]{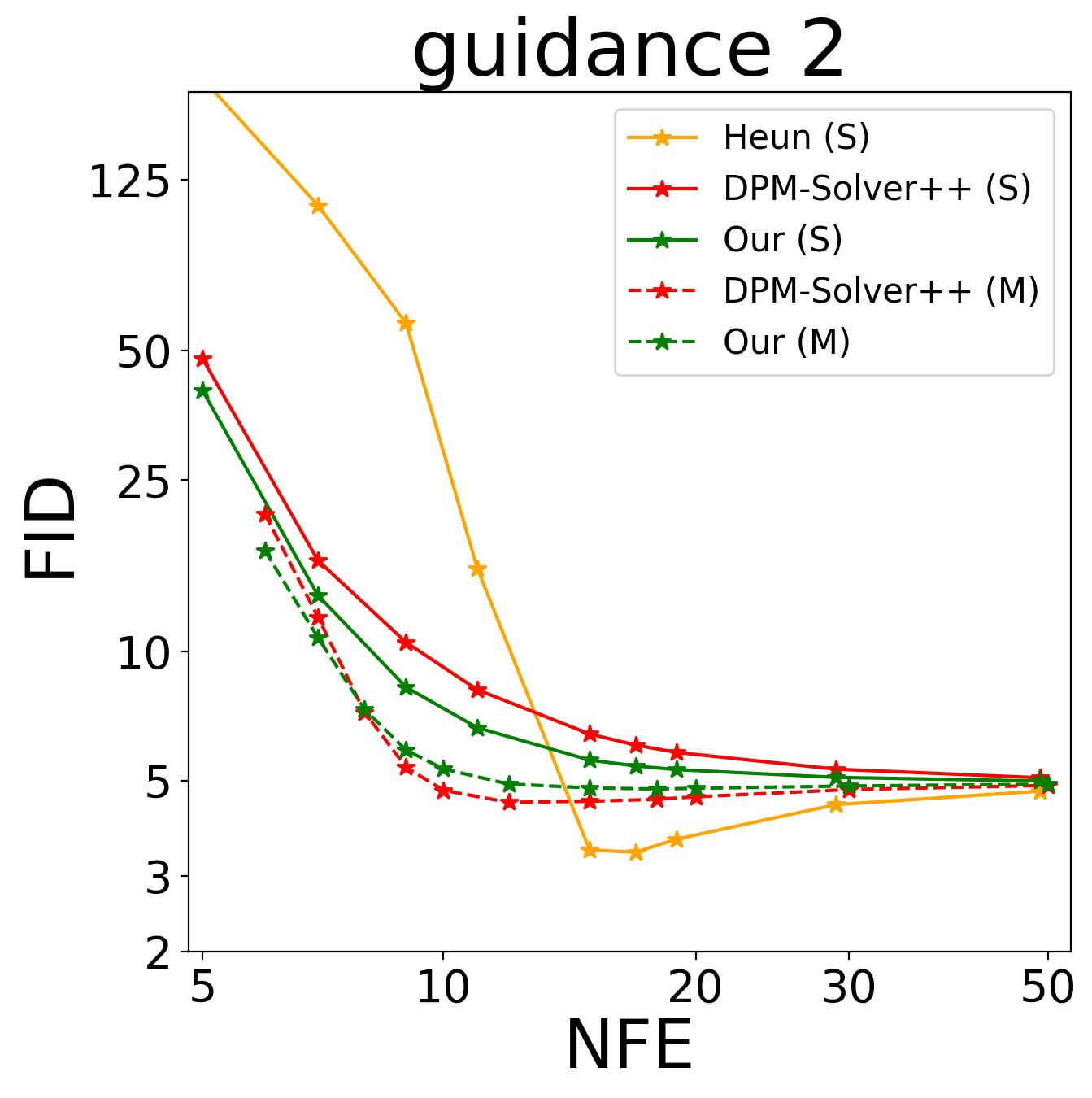}
\includegraphics[width=0.24\textwidth, height=0.22\textwidth, trim={0 1.5cm 0 0},clip]{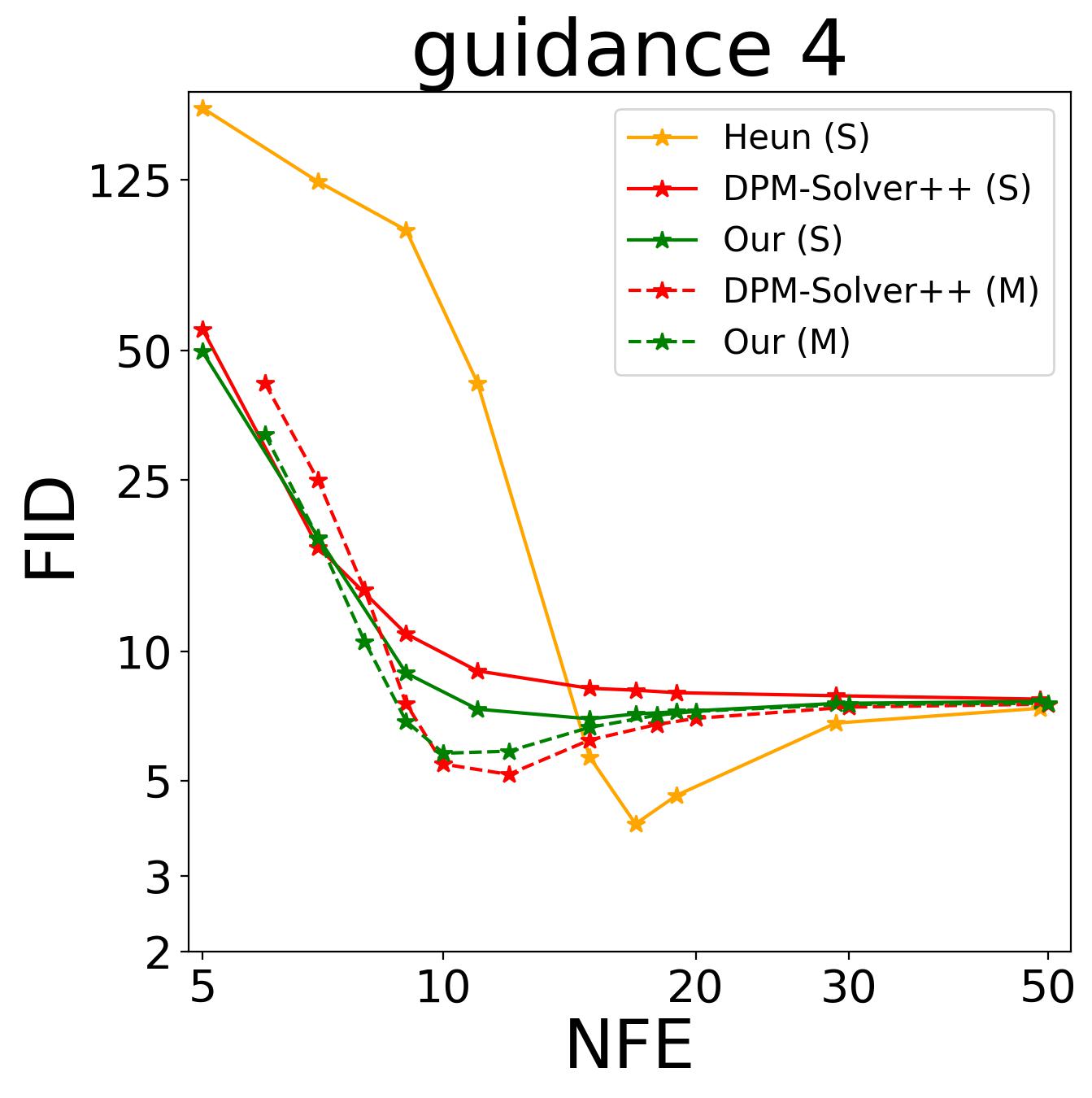}
\includegraphics[width=0.24\textwidth, height=0.22\textwidth, trim={0 1.5cm 0 0},clip]{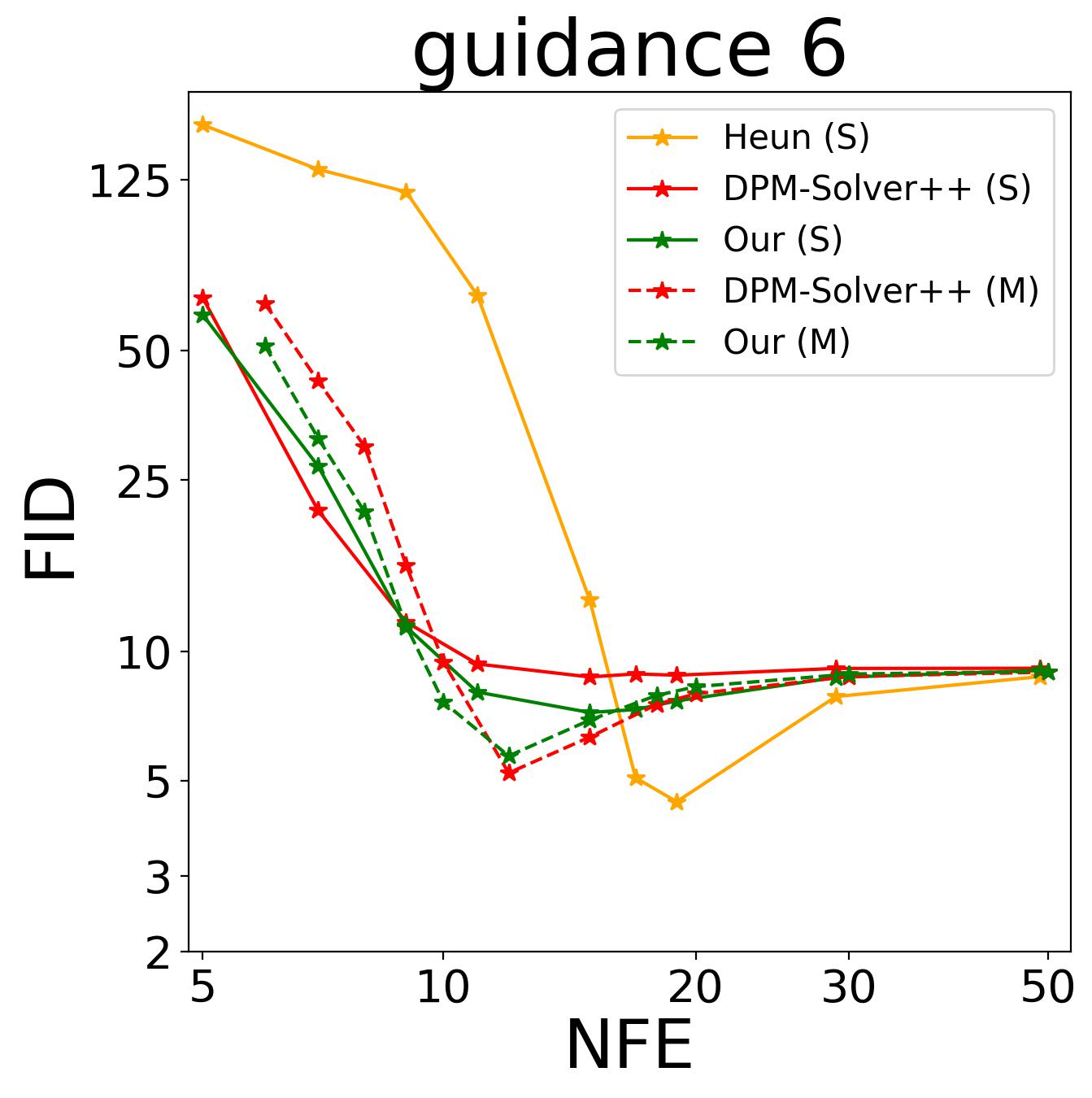}
   \\
   \includegraphics[width=0.24\textwidth, height=0.23\textwidth]{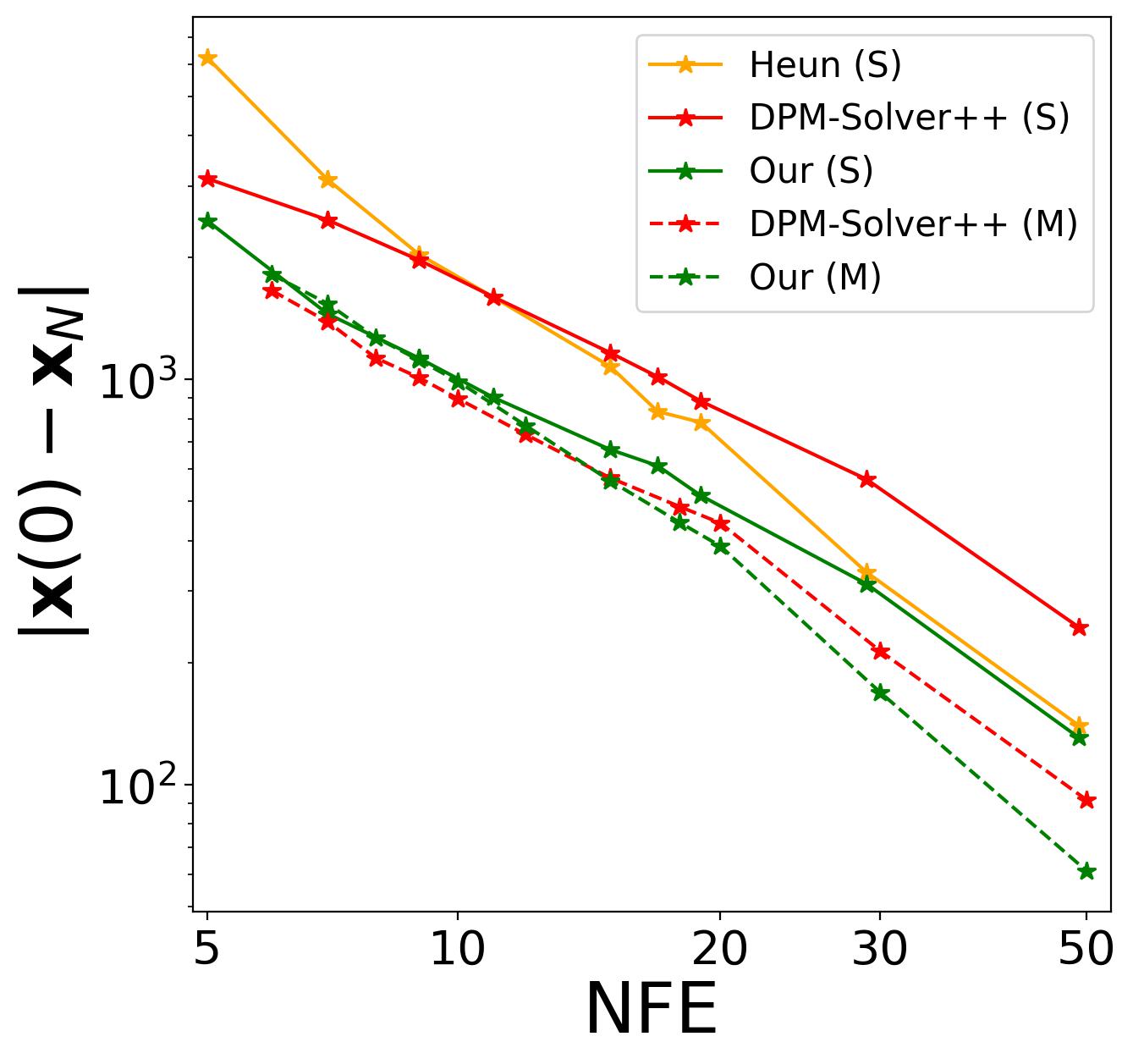}
   \includegraphics[width=0.24\textwidth, height=0.23\textwidth]{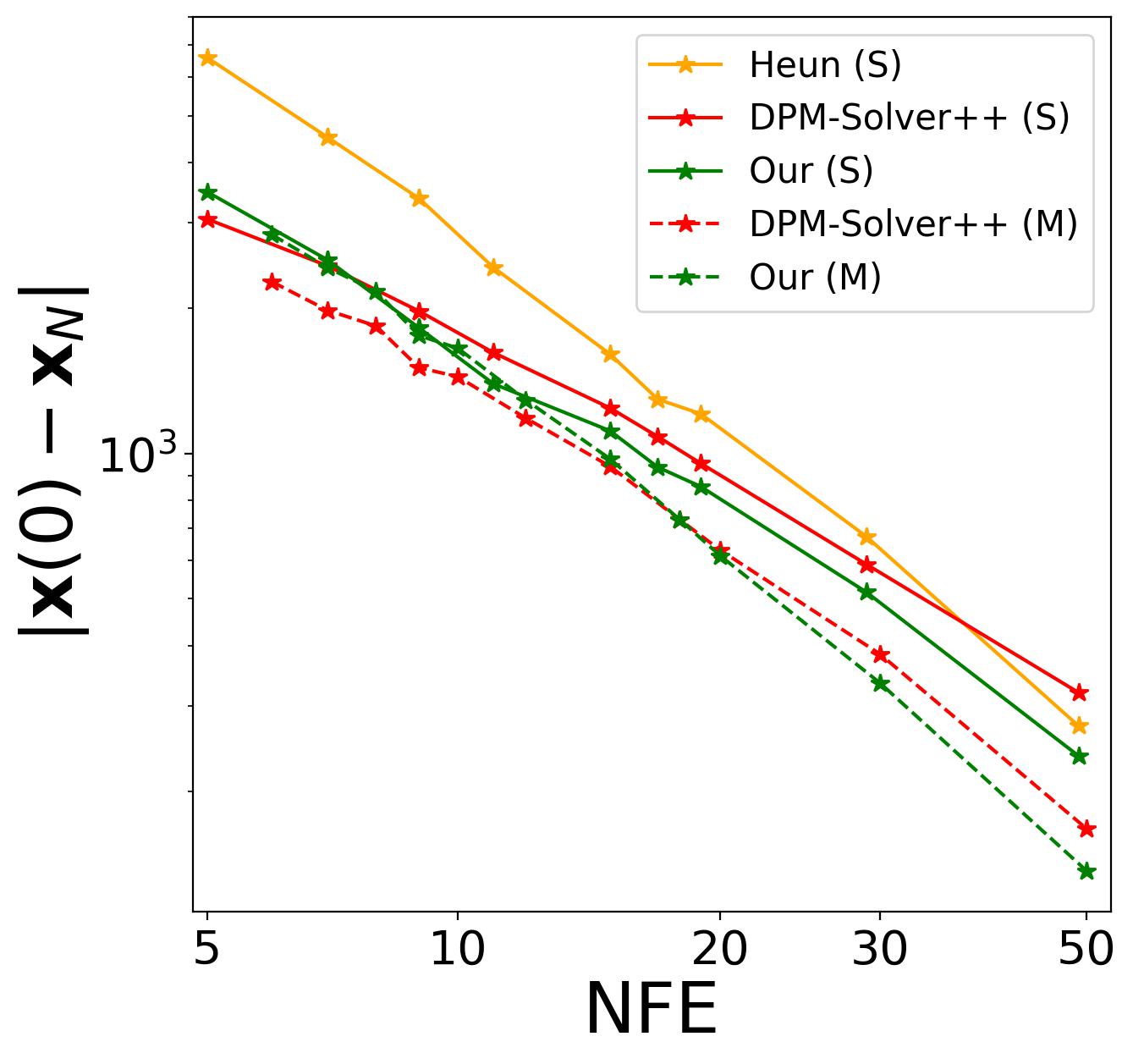}
   \includegraphics[width=0.24\textwidth, height=0.23\textwidth]{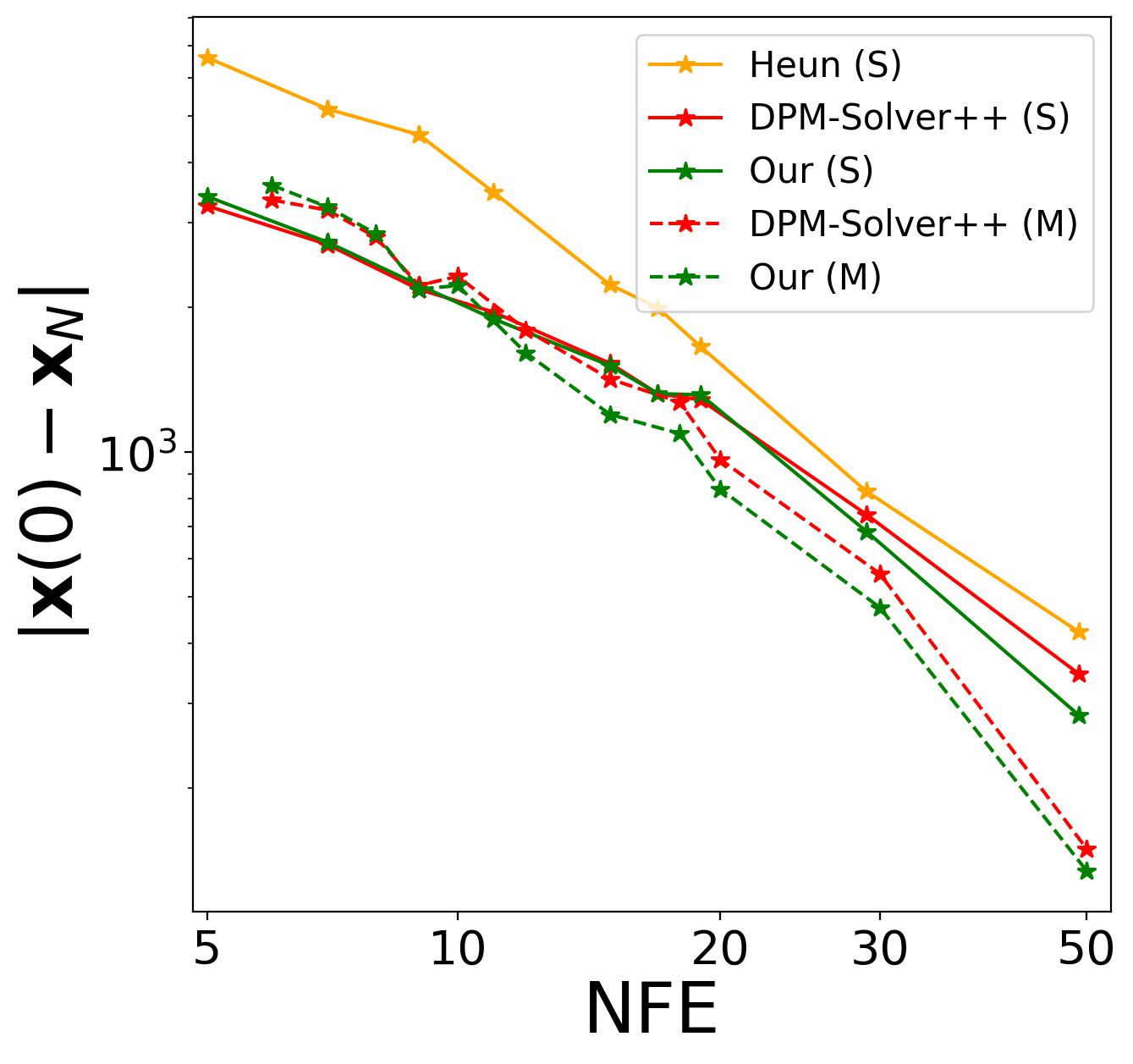}
   \includegraphics[width=0.24\textwidth, height=0.23\textwidth]{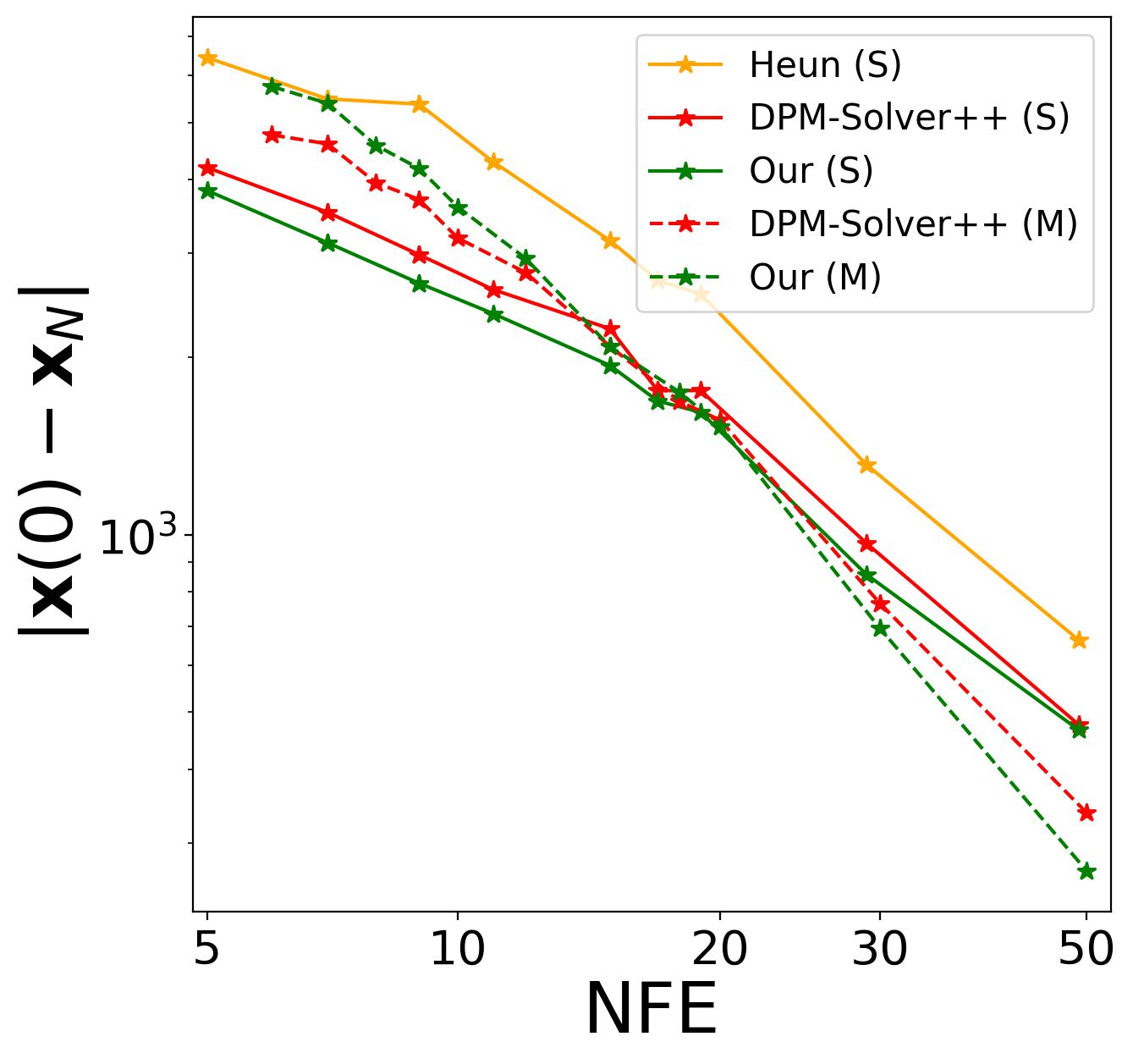}
   \caption{
    Comparison on DDPM-like classifier-guided diffusion models~\cite{dhariwal2021diffusion} with different guidance.
   }
   \label{fig:openai_guided_dm}
\end{figure}
We further investigate \our diffusion models trained using different schemes, including DDPM-like diffusion models with classifier-based guidance~\cite{dhariwal2021diffusion}. 
It is important to note that we employ dynamic thresholding~\cite{saharia2022photorealistic} for all experiments to help alleviate pixel over-saturation issues. To ensure a fair comparison, all algorithms differ solely in their numerical update schemes. As depicted in~\cref{fig:openai_guided_dm}, our previous observations remain valid for guided diffusion models trained with different techniques, and \our converges faster than baselines.

\begin{figure}
    \begin{subfigure}[b]{0.50\textwidth}
        \includegraphics[width=0.49\textwidth]{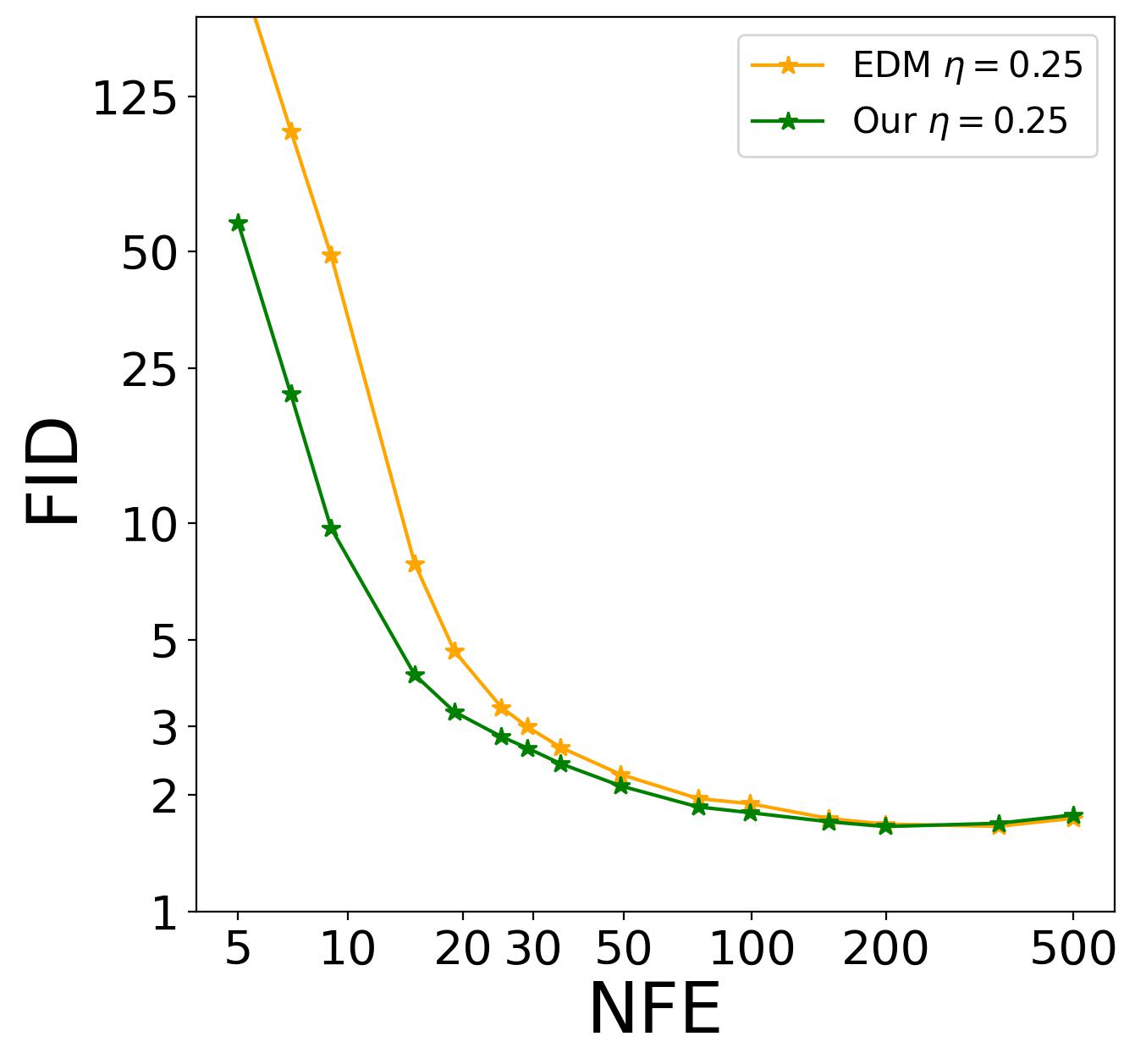}
        \includegraphics[width=0.49\textwidth]{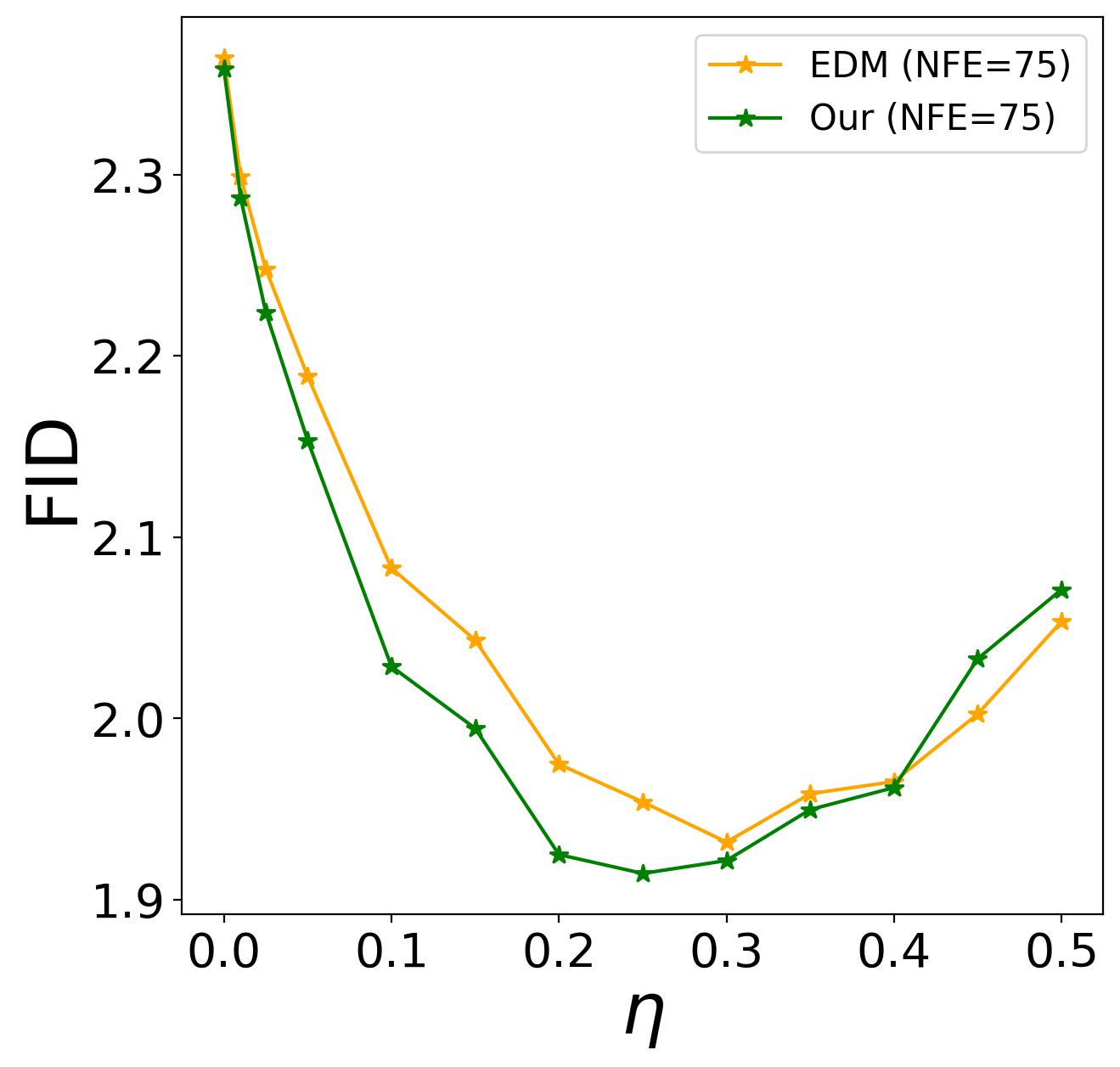}
        \caption{Stochastic sampling.}
        \label{fig:exp_stochastic_sampler}
    \end{subfigure}
    \begin{subfigure}[b]{0.50\textwidth}
        \includegraphics[width=0.49\textwidth]{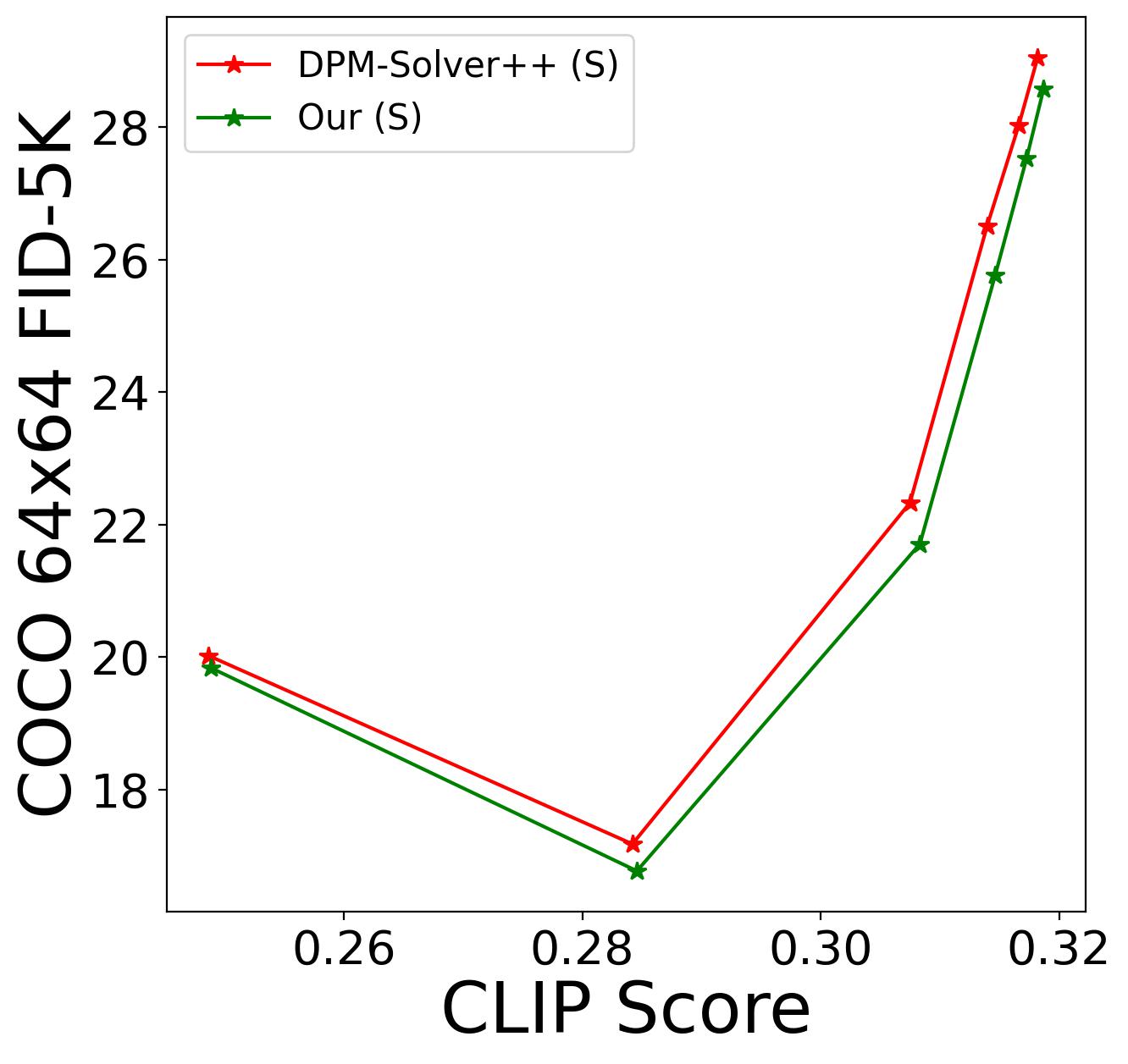}
        \includegraphics[width=0.49\textwidth]{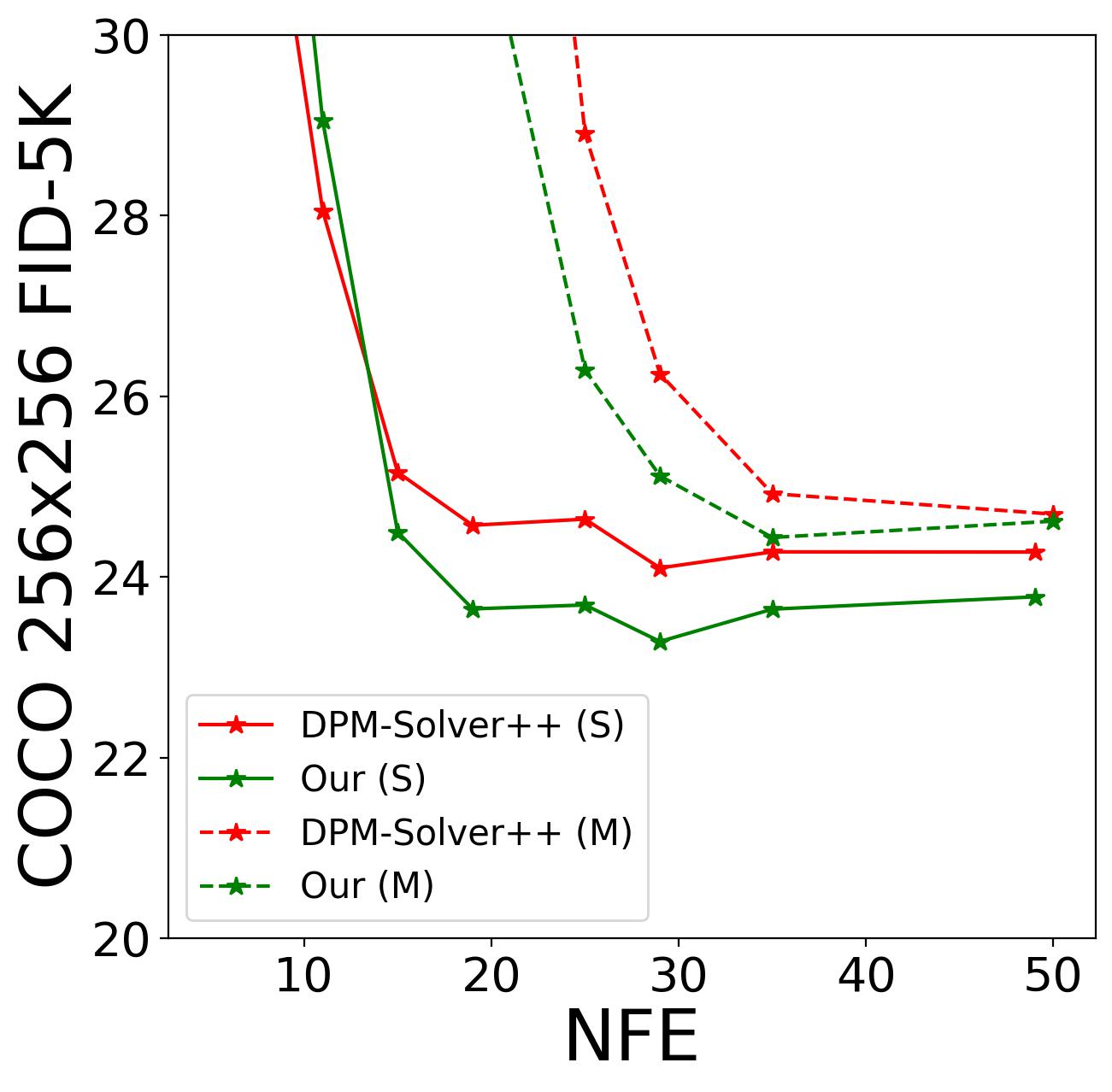}
        \caption{Comparison on DeepFloyd IF~\cite{deepfloyd23}}
        \label{fig:dfif}
    \end{subfigure}
    \caption{
        Our efficient single-step method can improve stochastic sampling as shown in~\cref{fig:exp_stochastic_sampler}. \our outperforms DPM-Solver++ in pixel space text-to-image diffusion models.
    }
    % \caption{Images A, B, and C.}
    \label{fig:}
\end{figure}

\paragraph{Stochastic sampling with improved single-step algorithm}
Existing works demonstrate that stochastic sampling can generate better quality results but demands significantly more computational resources. For example, EDM~\cite{karras2022elucidating} requires 511 NFEs. Encouraged by the improvement of single-step sampling, we conduct experiments to show that this enhancement can boost stochastic samplers and achieve a favorable trade-off between sampling quality and speed. First, we investigate how FID is affected by NFE for a given $\eta$. Next, we sweep different values of $\eta$ under the same NFE. As illustrated in~\cref{fig:exp_stochastic_sampler}, we can accelerate the stochastic sampler compared with Heun-based EDM sampler~\cite{karras2022elucidating}.

% \begin{figure}[!htp]
%     \centering
%     \includegraphics[width=4cm]{figure/plt/edm_imagenet_stoc_nfe.jpg}
%     \includegraphics[width=4cm]{figure/plt/edm_imagenet_stoc_eta.jpg}
%     \caption{Our efficient single-step method also improves stochastic sampling. }
%     \label{fig:exp_stochastic_sampler}
% \end{figure}

\paragraph{Cascaded text-to-image model} Finally, we test our sampling algorithm on cascaded text-to-image models based on DeepFloyd IF diffusion models~\cite{deepfloyd23}. We initially evaluate various sampling algorithms for $64\times 64$ tasks. We examine how image quality changes with an increasing number of NFEs in~\cref{fig:dfif}. Furthermore, we compare the FID-CLIP score trade-off curves for various classifier-free guidance schemes under the same computational budget. Next, we incorporate a super-resolution diffusion model conditioned on generated images from the low-resolution diffusion model. In various scenarios, we find that our model consistently outperforms DPM-Solver++ in terms of quality.

\section{Conclusion and limitations}

In this work, we present the Refined Exponential Solver~(\our), derived from a meticulous analysis of numerical defects in diffusion model probability flow ODEs and fulfilling all order conditions. \our enjoys superior error bounds theoretically and enhanced sampling efficiency in practice.
Originally designed to boost single-step updates, we have also extended the improvements to multistep schemes and stochastic samplers, effectively enhancing their performance.

Our work is also constrained by the following limitations.
Our analysis reveals that the performance of different samplers is contingent on the trajectory of network evaluations along the exact solution, where a time-smooth trajectory facilitates the reduction of numerical error. However, existing works are still uncertain about which time transformation or time scheduling leads to the most favorable trajectory. The logarithmic transformation we've used was selected based on our empirical observations.
In practical terms, training-free methods still require more than 10 network evaluations, making them significantly slower than GANs or distillation-based methods.
We leave it for future work to explore how distillation training can benefit from these improved training-free methods.

\clearpage
\bibliographystyle{apalike}
\bibliography{refs,add_refs}
\clearpage
\appendix

\section{Related works and discussion}~\label{app:related}

% \yc{is this sentence logically correct? It is good thus we need to accelerate it? Given the impressive generative capacity of diffusion models, a significant amount of research has been dedicated to accelerating these models.} 
% \qsh{
To unlock and democratize the extraordinary generative potential of diffusion models, considerable research efforts have been dedicated to enhancing their efficiency and speeding upsampling.
In addition to the works discussed in~\cref{sec:intro}, other methods have been explored to enhance the speed of diffusion models.

The authors of \citep{BaoLiZhu22} optimize the backward Markovian process to approximate the non-Markovian forward process, yielding an analytic expression of optimal variance in the denoising process. Another strategy involves making the forward diffusion nonlinear and trainable \citep{zhang2021diffusion,vargas2021solving,de2021diffusion,wang2021deep,chen2021likelihood}, in the spirit of the Schr\"odinger bridge \citep{chen2021stochastic}. However, this approach incurs a considerable training overhead.
Researchers have also explored modifying the backward stochastic process by incorporating more function evaluations and optimizing time scheduling~\citep{kong2021fast, watson2021learning}. Nevertheless, such acceleration strategies struggle to generate high-quality samples with a limited number of discretization steps. An alternate approach for improving diffusion models involves designing non-isotropic diffusion models, such as Blurring diffusion models~\citep{hoogeboom2022blurring,rissanen2022generative} and the critically-damped Langevin diffusion models~\citep{dockhorn2021score}.
There are other fast sampling methods by construction diffusion models in  latent space~\cite{rombach2022high,vahdat2021score}. In addition, several works show GANs can be leveraged to accelerate diffusion models~\citep{xiao2022DDGAN,wang2022diffusion}.

More related are the diffusion model sampling algorithms inspired by semilinear ODEs solvers~\citep{zhang2022fast,lu2022dpm}.
\citet{zhang2022fast} introduced Diffusion Exponential Integrator Sampler~(DEIS), whose fastest variant is based on an approximate exact solution, achieved by replacing the nonlinear function with polynomial exploration. 
However, DEIS was originally designed for fitting polynomials in the original $t$ space and is affected by noising scheduling $\sigma(t)$.
On the other hand, \citet{lu2022dpm} proposed DPM-Solver, which leverages a similar analysis of the exponential integrator to solve a semilinear ODE with time-varying coefficients. 
They further advanced DPM-Solver++ for the data prediction model and a sampling algorithm based on a multistep scheme, claiming improved sampling speed~\citep{lu2022dpmp}.
However, as we highlight in~\cref{ssec:order-condition}, their numerical scheme does not meet all the necessary order conditions. The omission of these conditions could potentially worsen performance, especially with non-isotropic ODEs~\citep{hochbruck2010exponential}. 
% \yc{who has studied the influence of order conditions for the following?} \qsh{Delete the following sentence to avoid confusion.}
% Such cases are explored in the context of non-isotropic diffusion models~\citep{hoogeboom2022blurring,dockhorn2021score}.
\section{Proof}
\label{app:proof}

In this section, we provide detailed derivations and proofs for the key theoretical results presented in the paper. Given that we have two semi-linear ODEs~\cref{eq:semi_ode,eq:semi_ode_eps} sharing similar structures, we primarily concentrate on the data prediction model, with the understanding that the results can be easily extended to the ODE based on the noise prediction model. Recall that we denote the denoiser network as $g$, and its evaluation along the exact solution $\vx(\lambda)$ as $f(\lambda) := g(\vx(\lambda),\lambda)$.

\subsection{Proof of~\Cref{lemma:intermediate-defect}}
\label{app:lemma:intermediate-defect}

% Now we start to prove~\Cref{lemma:intermediate-defect}. 
We first expand $f$ into a Taylor series with the remainder in integral form,

\begin{equation}~\label{eq:f-taylor}
    f(\lambda_n + \tau) = 
        \sum_{j=1}^q \frac{\tau^{j-1}}{(j-1)!} f^{(j-1)}(\lambda_n) + 
        \int_0^{\tau} \frac{(\tau - \nu)^{q-1}}{(q-1)!} f^{(q)}(\lambda_n + \nu) \diff \nu
\end{equation}
% \yc{where is the other hand? On the one hand,}\qsh{fixed now}
With~\cref{eq:f-taylor}, we can rewrite exact ODE solution~\cref{eq:voc} as 
\begin{align}~\label{eq:exact-solution-taylor}
    \vx(\lambda_n + c_i h) &= e^{-c_i h} \vx(\lambda_n) + \sum_{j=1}^{q_i} (c_i h)^j \phi_j(-c_i h) f^{(j-1)}(\lambda_n) 
     \\
     &+ \int_0^{c_i h} e^{-(c_i h - \tau)} \int_0^{\tau} \frac{(\tau - \nu)^{q_i-1}}{(q_i-1)!} f^{(q_i)}(\lambda_n + \nu) \diff \nu \diff \tau \nonumber,
\end{align}
where function $\phi_j$ is defined as
\begin{equation}\label{eq:def-phi}
    \phi_{j}(-h) = \frac{1}{h^j} \int_{0}^h e^{\tau - h} \frac{\tau^{j-1}}{(j -1)!} \diff \tau, \quad j \geq 1.
\end{equation}
With integration by part, we arrive at the recursion formulation for $\phi$ function~\cite{hochbruck2010exponential} as
\begin{equation}
    \phi_{k+1}(z)= 
\begin{cases}
    \frac{\phi_k(z) - 1 / k!}{z} &\quad k \geq 0, z \neq 0 \\
    e^z & \quad k = 0\\
    \frac{1}{k!}, &\quad z = 0, k \leq 0
\end{cases}.
\end{equation}

% On the other hand, by plugging~\cref{eq:f-taylor} into the numerical scheme~\cref{eq:sin-scheme}, we are able to derive another expression for exact solution.

% \begin{subequations}~\label{eq:app-int-defect}
%     \begin{align}
%         \vx(\lambda_n + c_i h) &= e^{-c_i h } \vx(\lambda_n) + h \sum_{j=1}^{i-1} a_{ij}(-h ) f(\lambda_n + c_j h) + \Delta _{ni} \\
%         \vx(\lambda_{n+1}) &= e^{-h } \vx(\lambda_n) + h \sum_{i=1}^s b_i(-h ) f(\lambda_n + c_i h) + \delta_{n+1}
%     \end{align}
% \end{subequations}
% where $\Delta _{ni}$ and $\delta_{n+1}$ are the error terms for intermediate states and final state defined in~\cref{eq:single-step-intermediate,eq:single-step-final}.

% Now by comparising exact solution $\vx(\lambda)$ in~\cref{eq:app-int-defect} and \cref{eq:exact-solution-taylor} at time $\lambda_n + c_i h$ and $\lambda_n + h$, we can reformulate the defects by
% \begin
One the other hand, by integrating~\cref{eq:f-taylor} into the numerical scheme~\cref{eq:sin-scheme}, we can derive another expression for the exact solution. We begin with the exact solution at the intermediate time $\lambda_n + c_i h$.
Plugging in~\cref{eq:f-taylor} into~\cref{eq:single-step-intermediate}, we obtain
\begin{align}~\label{eq:exact-solution-taylor-single-step-intermediate}
    \vx(\lambda_n + c_i h ) &= e^{-c_i h } \vx(\lambda_n) +
     h \sum_{k=1}^{i-1} a_{ik}(-h ) 
     \sum_{j=1}^{q_i} \frac{(c_k h)^{j-1}}{(j-1)!}f^{(j-1)}(\lambda_n) 
     \\
     & +
     h \sum_{k=1}^{i-1} a_{ik}(-h ) 
     \int_0^{c_k h} \frac{(c_k h - \nu)^{q_i-1}}{(q_i-1)!} f^{(q_i)}(\lambda_n + \nu) \diff \nu \diff \tau
     + \delta_{n,i}. \nonumber
\end{align}

By comparing~\cref{eq:exact-solution-taylor} and~\cref{eq:exact-solution-taylor-single-step-intermediate}, we obtain the error term $\Delta _{ni}$ as
\begin{align}
    \delta_{n,i} &= \sum_{j=1}^{q_i} h^j \psi_{j,i}(-h) f^{(j-1)}(\lambda_n) + \delta_{n,i}^{(q_i)} \\
    \psi_{j,i}(-h) &= \phi_j(-c_i h) c_i^j - \sum_{k=1}^{i-1} a_{ik}(-h) \frac{c_k^{j-1}}{(j-1)!},
\end{align}
where $\delta_{n,i}^{(q_i)}$ denotes higher order terms that resulted from the truncation of the Taylor series.

Similarly, we can get the expression for $\delta_{n+1}$ if we 
plug~\cref{eq:f-taylor} into~\cref{eq:single-step-final} and comparing the result against~\cref{eq:exact-solution-taylor}
\begin{align}
    \delta_{n+1} &= \sum_{j=1}^{q} h^j \psi_{j}(-h) f^{(j-1)}(\lambda_n) + \delta_{n+1}^{(q)} \\
    \psi_j(-h) &= \phi_j(-h) - \sum_{k=1}^s b_k(-h) \frac{c_k^{j-1}}{(j-1)!}
\end{align}
where $\delta_{n+1}^{(q)}$ denotes higher order terms that resulted from the truncation of the Taylor series.

\section{Proof of error bound}

We first state several mild assumptions that are required for our theoretical bounds. 

\begin{asum}~\label{asum:lipschitz}
    For nonlinear function $g: \mR^d \times \mR \rightarrow \mR^d$ considered in this work,
    \textit{e.g.} $\epsilon_\theta$, $\mD_\theta$, we assume there exists a real number $L(R)$ such that 
    % \yc{notation $\|\cdot\|$ instead of $|\cdot|$}\qsh{fixed now}
    \begin{equation}
        \norm{\vg(\vx_1, \lambda) - g(\vx_2, \lambda)} \leq L \norm{\vx_1 - \vx_2 }.
    \end{equation}
    for all $\lambda_{\min} \leq \lambda \leq \lambda_{max}$ and 
    $\max(\|\vx_1 - \vx(\lambda)\|, \|\vx_2 - \vx(\lambda)\|) \leq R$
    % \yc{$\max(\|\vx_1 - \vx(\lambda)\|, \|\vx_2 - \vx(\lambda)\|) \leq R$, does this make sense? $\vx(\lambda)$ is not unique without specifying initial condition} \qsh{I copy and paste from other paper, }
    where $\vx(\lambda)$ is one ODE solution with nonlinear function $g$.
\end{asum}
The region where $\vx_1, \vx_2$ exists and is close to the exact solution is referred to as the strip along the exact ODE solution. 
For high-order methods, we introduce another assumption regarding the nonlinear function $g$:
\begin{asum}
    For ODE with nonlinear function $g: \mR^d \times \mR \rightarrow \mR^d$ considered in this work,
    \textit{e.g.} $\epsilon_\theta$, $\mD_\theta$, $g$ is differentiable in a strip along the ODE exact solution $\vx(\lambda)$. All occurring derivatives are uniformly bounded.
    % \yc{this assumption includes the previous one?} \qsh{Oh, yes.}
\end{asum}
By default, we assume that~\cref{thm:order1,thm:order2} and~\cref{prop:order3} satiesfy above assumptions.
% \yc{why? By default, ~\cref{thm:order1,thm:order2} and~\cref{prop:order3} satiesfy above assumptions.} 
First, we consider a scenario with a uniform step size, denoted as $h$. Subsequently, the obtained results are extended to accommodate non-uniform step sizes. This generalization involves loosening the error bounds' reliance on the step size from the uniform case, directing it instead towards the maximum step size evident in the non-uniform scenario. Indeed, the error boundaries delineated in ~\cref{thm:order1,thm:order2} manifest dependence on this longest step size.

First, we bound the truncated high-order term $\delta_{n,i}^{(q_i)}$. Based on~\cref{eq:exact-solution-taylor,eq:exact-solution-taylor-single-step-intermediate}, $\delta_{n,i}^{(q_i)}$ can be bounded by 
% \yc{notation $\|\cdot\|$ instead of $|\cdot|$ throughout}
\begin{align}
    \|\delta_{n,i}^{(q_i)}\| \leq 
    & \|\int_0^{c_i h} e^{-(c_i h - \tau)} \int_0^{\tau} \frac{(\tau - \nu)^{q_i-1}}{(q_i-1)!} f^{(q_i)}(\lambda_n + \nu) \diff \nu \diff \tau\| \nonumber\\
    &+ \|h \sum_{k=1}^{i-1} a_{ik}(-h ) 
    \int_0^{c_k h} \frac{(c_k h - \nu)^{q_i-1}}{(q_i-1)!} f^{(q_i)}(\lambda_n + \nu) \diff \nu\| \nonumber \\
    & \leq \sup_{\lambda \in [\lambda_n , \lambda_n + c_i h]} \|f^{(q_i)}\| \left(\| 
    \int_0^{c_i h} e^{-(c_i h - \tau)} \int_0^{\tau} \frac{(\tau - \nu)^{q_i-1}}{(q_i-1)!} \diff \nu \diff \tau 
    \| \right. \nonumber\\
    & \left. \quad + \|
    h \sum_{k=1}^{i-1} a_{ik}(-h ) 
    \int_0^{c_k h} \frac{(c_k h - \nu)^{q_i-1}}{(q_i-1)!} \diff \nu
    \|
    \right) \nonumber \\
    & \leq \sup_{\lambda \in [\lambda_n , \lambda_n + c_i h]} \|f^{(q_i)}\|C (c_i h)^{q_i + 1} \label{app-eq:delta_ni_bound},
\end{align}
where $C$ is a sufficiently large constant. 
% \yc{something is miss. what if $\|f^{(q_i)}\|=0$? the bound is 0? It is not clear why the second term below, that is independent of $\|f^{(q_i)}\|$, can be bounded by $\sup_{\lambda \in [\lambda_n , \lambda_n + c_i h]} \|f^{(q_i)}\|C (c_i h)^{q_i + 1}$ that is proportional to $\|f^{(q_i)}\|$}
This is due to
\begin{align*}
    \|\int_0^{c_i h} e^{-(c_i h - \tau)} \int_0^{\tau} \frac{(\tau - \nu)^{q_i-1}}{(q_i-1)!} \diff \nu \diff \tau \| \leq
        \|\int_0^{c_i h} \diff \tau\| \| \int_0^{c_i h } \frac{(c_i h - \nu)^{q_i-1}}{(q_i-1)!} \| = \frac{(c_i h)^{q_i + 1}}{(q_i)!} \\
    \|
    h \sum_{k=1}^{i-1} a_{ik}(-h ) 
    \int_0^{c_k h} \frac{(c_k h - \nu)^{q_i-1}}{(q_i-1)!} \diff \nu
    \|
    \leq \frac{\|\sum_{k=1}^{i-1} a_{ik}(-h )\|}{c_i} \frac{(c_i h)^{q_i + 1}}{(q_i)!}.
\end{align*}
% \yc{how to estimate $C$? where does $|
%     h \sum_{k=1}^{i-1} a_{ik}(-h ) 
%     \int_0^{c_k h} \frac{(c_k h - \nu)^{q_i-1}}{(q_i-1)!} \diff \nu
%     |$ go?}
We note above conclusion only holds when $\psi_{j,i}(-h) = 0$ is satisfied. 
% \yc{$\phi_{j,i}(-h)$ is not defined}\qsh{fixed}
Otherwise, defects of $h^j f^{(j-1)}(\lambda_n)$ will show up $\|\delta_{n,i}^{(q_i)}\|$ once $\psi_{j,i}(-h) \neq 0$.

Similarly, we can bound the accumulation of $\delta_i^{(q_i)}$ with 
% \yc{what does the superscript $(r)$ mean below?}\qsh{fixed now}
\begin{equation}~\label{eq:acc_bound}
    \| \sum_{j=0}^{n-1} e^{-j h} \delta_{n-j}^{(q_i)}\| \leq C h^{q_i} \sup_{\lambda \in [\lambda_1 , \lambda_n]} \|f^{(q_i)}\|
\end{equation}
once $\phi_j = 0$ is satisfied. Indeed,
the defect $\delta_{n}^{(q)}$ can be formulated
\begin{align}
    \|\delta_{n}^{(q_i)}\| &= \|\int_{0}^h e^{-(h - \tau)} \int_{0}^\tau \frac{(\tau - \nu)^{(q_i-1)}}{(q_i -1)!} f^{(q_i)}(\lambda_{n-1} + \nu) \diff \nu \diff \tau \| \\
    & \leq \|\int_0^h e^{-(h - \tau)} \diff \tau\| \| \int_0^h \frac{(h -\nu)^{(q_i-1)}}{(q_i -1)!}\diff \nu\| \sup_{\lambda \in [\lambda_{n-1} , \lambda_n ]} \|f^{(q_i)}\| \\
    & = (1-e^{-h})\frac{h^{q_i}}{q_i !} \sup_{\lambda \in [\lambda_{n-1} , \lambda_n ]} \|f^{(q_i)}\|
\end{align}
Therefore, we can bound
\begin{align*}
    \| \sum_{j=0}^{n-1} e^{-j h} \delta_{n-j}^{(q_i)}\| &\leq \sum_{j=0}^{n-1} \|e^{-j h} \delta_{n-j}^{(q_i)}\|  \\
    & \leq \|\sum_{j=0}^{n-1} e^{-jh}\| (1-e^{-h})\frac{h^{q_i}}{q_i !} \sup_{\lambda \in [\lambda_{1} , \lambda_n ]} \|f^{(q_i)}\| \\
    & \leq \frac{1}{1 -e^{-h}} (1-e^{-h})\frac{h^{q_i}}{q_i !} \sup_{\lambda \in [\lambda_{1} , \lambda_n ]} \|f^{(q_i)}\| \\
    & = \frac{h^{q_i}}{q_i !} \sup_{\lambda \in [\lambda_{1} , \lambda_n ]} \|f^{(q_i)}\|
    \leq C h^{q_i} \sup_{\lambda \in [\lambda_1 , \lambda_n]} \|f^{(q_i)}\|,
\end{align*}
with a sufficiently large $C$.

Second, we introduce Discrete Gronwall Inequality in~\Cref{lemma:dgi}, which we will use later.

\begin{lemma}[Discrete Gronwall Inequality]~\label{lemma:dgi} 
% \yc{present this earlier}
    Let $\langle \alpha_n \rangle$ and $\langle \beta_n \rangle$ be nonnegative sequences and $c$ a nonnegative constant. If
    \[
    \alpha_n \leq c + \sum_{0 \leq k < n} \beta_k \alpha_k,
    \]
    then
    \begin{equation}
        \alpha_n \leq c \prod_{0 \leq j < n} (1+\beta_j) \leq c \exp{(\sum_{0 \leq j < n} \beta_j)} \quad \text{for} \quad n \geq 0.
    \end{equation}
\end{lemma}

\subsection{Order $1$ error bound for~\cref{thm:order1}}

With~\Cref{lemma:intermediate-defect}, we can formulate the error recursion for the $s=1$ case
\begin{equation}
    \Delta_{n+1} = e^{-h} \Delta_n + h \phi_1(-h)(g(\vx_n, t_n) - f(t_n)) - \delta_{n+1}
\end{equation}
with defects $\delta_{n+1} = \delta_{n+1}^{(1)}$, the recursion gives
\begin{equation}
    \Delta_n = h\sum_{j=0}^{n-1} e^{(-n-j-1)h} \phi_1(-h) (g(\vx_j, t_j) - f(t_j)) - \sum_{j=0}^{n-1} e^{-j h} \delta_{n-j}. 
\end{equation}
Thanks to~\cref{asum:lipschitz} and~\cref{eq:acc_bound}, we can bound above $\Delta_n$ by
% \yc{is the following correct? where does the Lip-constant $L$ of $g$ show up? also, $q_i$ should be 1?} 
\begin{equation}~\label{app-eq:order1-explicit-delta-n}
    \norm{\Delta_n} \leq \sum_{j=1}^{n-1} h\phi_1(-h) e^{-(n-j-1)h} L \|\Delta_j\| + Ch \sup_{\lambda \in [\lambda_n , \lambda_n + c_i h]} \|f^{(1)}\|
\end{equation}

With Discrete Gronwall Inequality~\Cref{lemma:dgi} and $\Delta_0 =0$, we can show that
\begin{align}
    \norm{\Delta_n} = \norm{\vx_n - \vx(\lambda_n)} &\leq Ch \sup_{\lambda_{\text{min}} \leq \lambda \leq \lambda_\text{max}} \norm{f^\prime(\lambda)} \exp{(\prod_{j=1}^{n-1} hL\phi_1(-h) e^{-(n-j-1)h} )} \\
    & \leq Ch \sup_{\lambda_{\text{min}} \leq \lambda \leq \lambda_\text{max}} \norm{f^\prime(\lambda)} \exp(L^n) \\
    & \leq C' h \sup_{\lambda_{\text{min}} \leq \lambda \leq \lambda_\text{max}} \norm{f^\prime(\lambda)} \label{eq:app-order1-uni}
\end{align}
with a large enough $C'$.
This finishes the proof for~\cref{thm:order1} if we upper bound \cref{eq:app-order1-uni} by the largest step size in the non-uniform stepsize case. 
Thanks to order condition $\psi_1 =0$ and $c_1 = 0$, the Buther tableau follows
\begin{equation}
    \begin{array}
            {c|c}
            0 &  \\
            \hline
             & \phi_1(-h).
    \end{array}
\end{equation}

\subsection{High order error bounds}

We first start with the second order method.
Since it is a second-stage algorithm, we have only one intermediate point $\vx_{n,1}$. Based on \cref{eq:single-error-recursion,app-eq:delta_ni_bound} and \cref{asum:lipschitz}, we can bound
\begin{align}
    \norm{\Delta_{n,2}} &\leq C_1 \norm{\Delta_n} + \|\delta_{n,2}^{(1)}\| \\
                & \leq C_1 \norm{\Delta_n} + C_2 h^2 \sup_{\lambda \in [\lambda_n, \lambda_n + c_2 h]]} \|f^{(1)}\|,
\end{align}
where $C_1,C_2$ are two constants. And we know $\Delta_{n,1} = \Delta_n$ for our explicit method due to $c_1 = 0$.

Similar to~\cref{app-eq:order1-explicit-delta-n}, we can bound
\begin{align*}
    \norm{\Delta_n} &\leq h\sum_{j=0}^{n-1} e^{(-n-j-1)h} \phi_1(-h) C_3 \max{(\norm{\Delta_j}, \norm{\Delta_{j,1}})} + \ \sum_{j=0}^{n-1} e^{-j h} \delta_{n-j}\| \\
    & \leq h\sum_{j=0}^{n-1} e^{(-n-j-1)h} \phi_1(-h) C_4 \norm{\Delta_j} + h\sum_{j=0}^{n-1} e^{(-n-j-1)h} \phi_1(-h) C_2 h^2 \sup_{\lambda \in [\lambda_{min}, \lambda_{max}]} \|f^{(1)}\| \\
    & \quad + C_5 h^2 \sup_{\lambda \in [\lambda_{min}, \lambda_{max}]} \|f^{(2)}\|,
\end{align*}
where $C_4 = \max{(1,C_1)}$, and the last inequality is due to ~\cref{eq:acc_bound}.
We note that $h\sum_{j=0}^{n-1} e^{(-n-j-1)h} \phi_1(-h) \leq 1$. We can further simplify the upper bound of $\Delta_n$ by
\begin{align*}
    \norm{\Delta_n} \leq &h\sum_{j=0}^{n-1} e^{(-n-j-1)h} \phi_1(-h) C_4 \norm{\Delta_j} \\
    &+ \max{(C_2, C_5)}h^2(\sup_{\lambda \in [\lambda_{min}, \lambda_{max}]} \|f^{(2)}\| + \sup_{\lambda \in [\lambda_{min}, \lambda_{max}]} \|f^{(1)}\|).
\end{align*}
Due to~\Cref{lemma:dgi}, we can upper bound it by
\begin{align}
    \norm{\Delta_n} \leq &\max{(C_2, C_5)}h^2(\sup_{\lambda \in [\lambda_{min}, \lambda_{max}]} \|f^{(2)}\| + \sup_{\lambda \in [\lambda_{min}, \lambda_{max}]} \|f^{(1)}\|) \\
    & \quad \exp{(\prod_{j=1}^{n-1} C_4 h\phi_1(-h) e^{-(n-j-1)h} )}  \nonumber \\
    & \leq C' h^2 (\sup_{\lambda \in [\lambda_{min}, \lambda_{max}]} \|f^{(2)}\| + \sup_{\lambda \in [\lambda_{min}, \lambda_{max}]} \|f^{(1)}\|) \label{eq:app-order2-uni}
\end{align}
for a large enough $C'$. This finishes the proof of~\cref{thm:order2} if we upper bound \cref{eq:app-order1-uni} by the largest step size in the non-uniform stepsize case. 

Thanks to condition $\psi_1(-h) = \psi_2(-h) = \psi_{1,2}(-h) = 0$, the Buther tableaus has to satisfy
\begin{align*}
    b_1 + b_2  &= \phi_1(-h) \\
    b_2 c_2 &= \phi_2(-h) \\
    a_{21} &= c_2 \phi_1(-c_2h).
\end{align*}
Therefore, the only solution for Buther tableau is
\begin{equation}
    \begin{array}{c|cc}
                    0 \\
                    c_2 & c_2 \phi_1(-c_2h) \\
                    \hline
                    0 & \phi_1(-h) - \frac{1}{c_2} \phi_2(-h) & \frac{1}{c_2} \phi_2(-h)
                \end{array}.
\end{equation}

For third-order methods, numerical methods with three-stage methods have to satiesfiy~\citep[Sec 5.2]{hochbruck2005explicit}
\begin{align*}
    \psi_1(-h) &= 0 \\
    \psi_2(-h) &= 0 \\
    \psi_{1,2}(-h) &= 0 \\
    \psi_{1,3}(-h) &= 0 \\
    \psi_3(-h) &= 0 \\
    b_2 \psi_{2,2}(-h) + b_3 \psi_{2,3}(-h) &=0
\end{align*}

One two-parameter solution family for Buther tableau satisfies other conditions and achieves third-order error bounds follows
\begin{equation}
    \begin{array}
            {c|ccc}
            0\\
            c_2 & c_2 \phi_{1,2} \\
            c_3 & c_3 \phi_{1,3} - a_{32} & \frac{c_3^2}{c_2} \phi_{2,3} \\
            \hline
                & \phi_1 - b_2 - b_3 & 0 & \frac{1}{c_3} \phi_2
    \end{array},
\end{equation}
where $\gamma$ can be obtained by solving $2(\gamma c_2 + c_3) = 3(\gamma c_2^2 + c_3^3)$ once $c_2, c_3$ are given.

% \begin{equation}
%     \norm{\delta_{n,i}} \leq C \Delta_n + c h ^2 \sup_{\lambda \in [\lambda_n , \lambda_n + c_i h]} |f^{(q_i)}| 
%     \quad
%     \norm{\sum_{j=0}^{n-1} e^{-j h} \Delta_{n-j}^{(2)}} \leq C h^2 \sup_{\lambda \in [\lambda_n , \lambda_n + c_i h]} |f^{(q_i)}| 
% \end{equation}
\section{Algorithms}~\label{app:algo}

For deterministic single-step and stochastic samplers, we have listed the unified algorithm in~\cref{alg:eis-single}. 
We have also listed the algorithm for single-step second order update in~\cref{alg:eis-single-second}.
Furthermore, we list general high order algorithm in~\cref{alg:eis-single-high}.

\begin{algorithm}[t]
    \footnotesize
    \captionof{algorithm}[single-step]{\atphantom\ \our General $s$-stage Single-step Update with $\{c_s\}$}
    \begin{spacing}{1.1}
    \begin{algorithmic}[1]
      \AProcedure{SingleUpdateStep}{$\vx_n, \sigma_n, \sigma_{n+1}$}
            \AState{$\lambda_{n+1}, \lambda_n \gets -\log(\sigma_{n+1}),  -\log( \sigma_n)$}
            \AState{$h \gets \lambda_{n+1} - \lambda_n$}
                \AComment{Step length}
            \AState{$\{a_{ij}\}, \{b_j\} \gets$ Butcher tableau with $\{c_i\}$}
                \AComment{Single-step update coeffcients}
            \AState{$(\vx_{n,1},\,\lambda_{n,1}) \gets (\vx_n,\,\lambda_n)$}
                \AComment{$c_1=0$ for explicit methods}
            \AFor{$i$ in $2,\cdots,s$}
                \AState{$(\vx_{n,i},\,\lambda_{n,i}) \gets (e^{-c_i h} \vx_n + h \sum_{j=1}^i a_{i,j} \mD_\theta(\vx_{n,j}, \lambda_n + c_j h),\,\lambda_n + c_i h)$}
            \EndFor
          \AState{$\vx_{n+1} \gets e^{-h}\vx_n + h(\sum_{i=1}^s b_i \mD_\theta(\vx_{n,i}, \lambda_n + c_i h))$}
        \AState{\textbf{return} $\vx_{n+1}$}
      \EndProcedure
    \end{algorithmic}
    \end{spacing}
    \label{alg:eis-single-high}
\end{algorithm}

\subsection{Multistep algorithm}

The insights and analysis gained from a multi-stage single-step numerical scheme can be leveraged to develop multistep numerical methods. The crux of a multistep update step lies in its utilization of not only the function evaluation at the current state but also previous function evaluations.

To underscore the parallels between single-step and multistep methods, consider a single update step from timestamp $\lambda_n$ to $\lambda_{n+1}:=\lambda_n + h$ with function evaluations $\{ g(\vx_{n,i}, \lambda_n + c_i h)\}_{i=1}^r$, where $c_1=0, c_i < 0$ for $1 < i \leq r$ and $\vx_{n,i}$ are numerical results on timestamp $\lambda_n +c_i h$. The general multistep scheme can be expressed as follows:
\begin{equation}
    \vx_{n+1} = e^{-h} \vx_n + h[\sum_{i=1}^r b_i g(\vx_{n,i}, \lambda_n + c_i h)]
\end{equation}
The present multistep method configuration bears many resemblances to the single-step method outlined in~\cref{ssec:ode-error-analysis}. The distinction resides in the selection of $\{c_i\}_{i=2}^r$, which are chosen such that $\lambda_n + c_i h = \lambda_{n+1 -i}$. Consequently, with $\vx_{n,i}$ coinciding with $\vx_{n+1-i}$, we can bypass the function evaluation cost, as $g(\vx_{n,i}, \lambda_{n} + c_i h)$ is readily available due to the existing value of $g(\vx_{n+1-i}, \lambda_{n+1 -i})$.

From this viewpoint, the construction of auxiliary $\hat{\vx}$ and intermediate numerical defects $\delta_n$ in~\cref{ssec:ode-error-analysis} remain applicable for multistep cases. Most notably, the error recursion in~\cref{eq:error-recursion-final} and the expansion of $\delta_{n+1}$ in~\Cref{lemma:intermediate-defect} persist in the multistep scenario. To minimize numerical defects, we also aim to reduce ${\delta_{n} }$. Therefore, the choice of ${b_i}$ from the Butcher tableaus developed for the single-step scheme can be employed for multistep methods with particular $c_i$ values. The second-order multistep algorithm is detailed in~\cref{alg:eis-multi-second}.

\begin{algorithm}[t]
    \footnotesize
    \captionof{algorithm}[single-step]{\atphantom\ \our Second order Multistep Update Scheme}
    \begin{spacing}{1.1}
    \begin{algorithmic}[1]
      \AProcedure{MultiStepUpdate}{$\vx_n, \sigma_n, \sigma_{n+1}, \mD_\theta(\vx_{n-1}, \lambda_{n-1})$ % \yc{the notation $i$ and $n$ are messed up}
      }
            \AState{$\lambda_{n+1}, \lambda_n, \lambda_{n-1} \gets -\log(\sigma_{n+1}),  -\log( \sigma_n),  -\log( \sigma_{n-1})$}
            \AState{$h \gets \lambda_{n+1} - \lambda_n$}
                \AComment{Step length}
            \AState{$c_2 \gets \frac{\lambda_{n-1} - \lambda_n}{h}$}
            \AState{$b_1, b_2 \gets \phi_1(-h) - \phi_2(-h) / c_2, \phi_2(-h) / c_2$}
                \AComment{Coeffcients}
          \AState{$\vx_{n+1} \gets e^{-h}\vx_n + h(b_1 D_\theta(\vx_n, \lambda_n) + b_2 D_\theta(\vx_{n-1}, \lambda_{n-1}))$}
        \AState{\textbf{return} $\vx_{n+1}$}
      \EndProcedure
    \end{algorithmic}
    \end{spacing}
    \label{alg:eis-multi-second}
\end{algorithm}

\section{Experiment details}\label{app:expr}

In this section, we provide necessary experiments details and extra experiments.

\subsection{Reproduction of \cref{fig:teaser}}

\paragraph{\cref{fig:function-derivative}} We use the pre-trained ImageNet $64\times 64$ class-conditioned diffusion model~\citep{karras2022elucidating} to conduct the experiment. 
Due to the lack of ground truth solution trajectories, we use high-accuracy ODE solvers to approximate the solutions.
Concretely, we employ RK4 with 500 steps~(2000 NFE) to generate trajectories, with negligible alterations with more steps. 
To generate those trajectories, we randomly select initial random noise and labels.
Curves in~\cref{fig:function-derivative} are averaged over $512$ trajectories.

\begin{figure}[!htp]
    \centering
    \includegraphics[width=4cm]{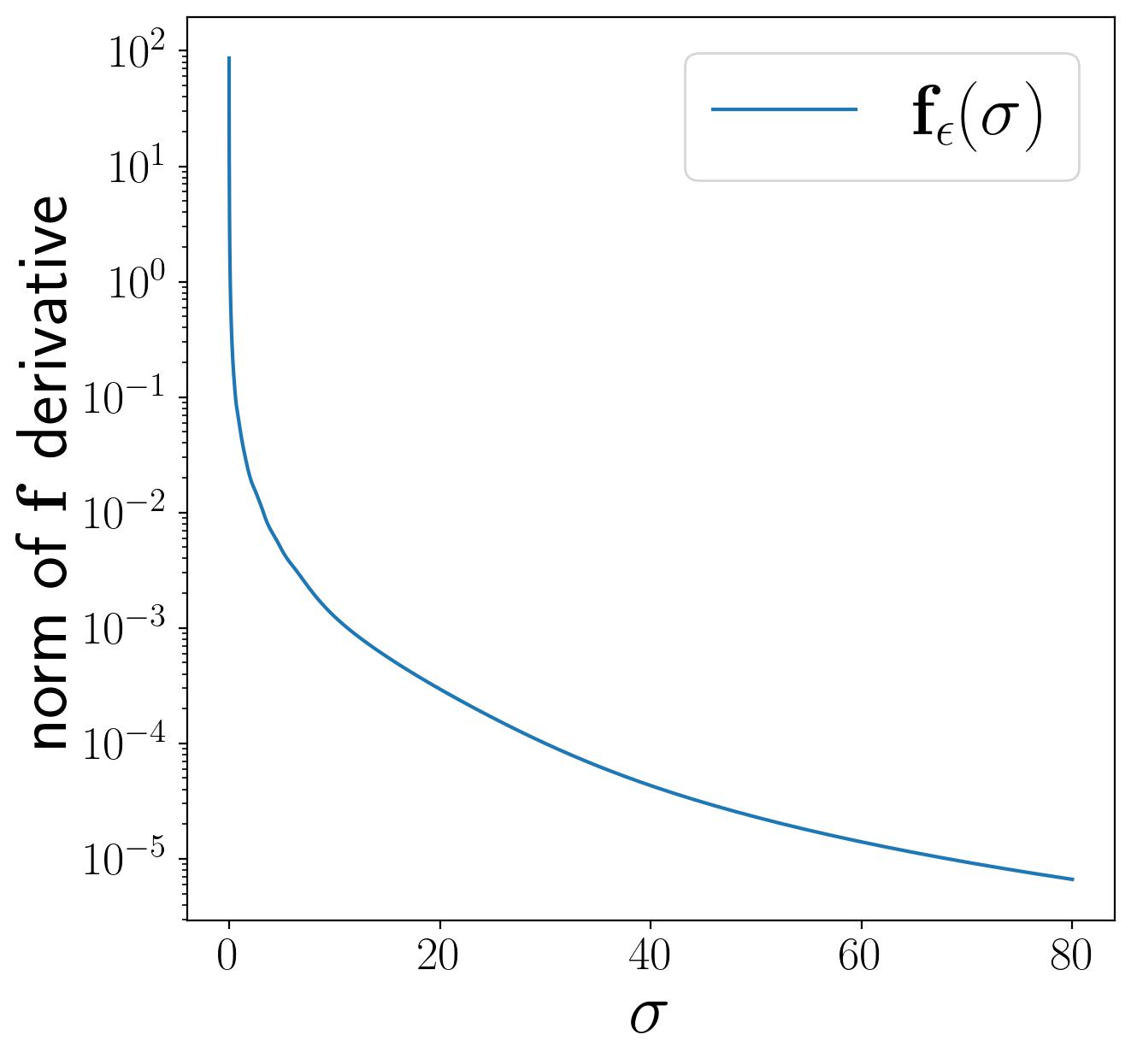}
    \includegraphics[width=4cm]{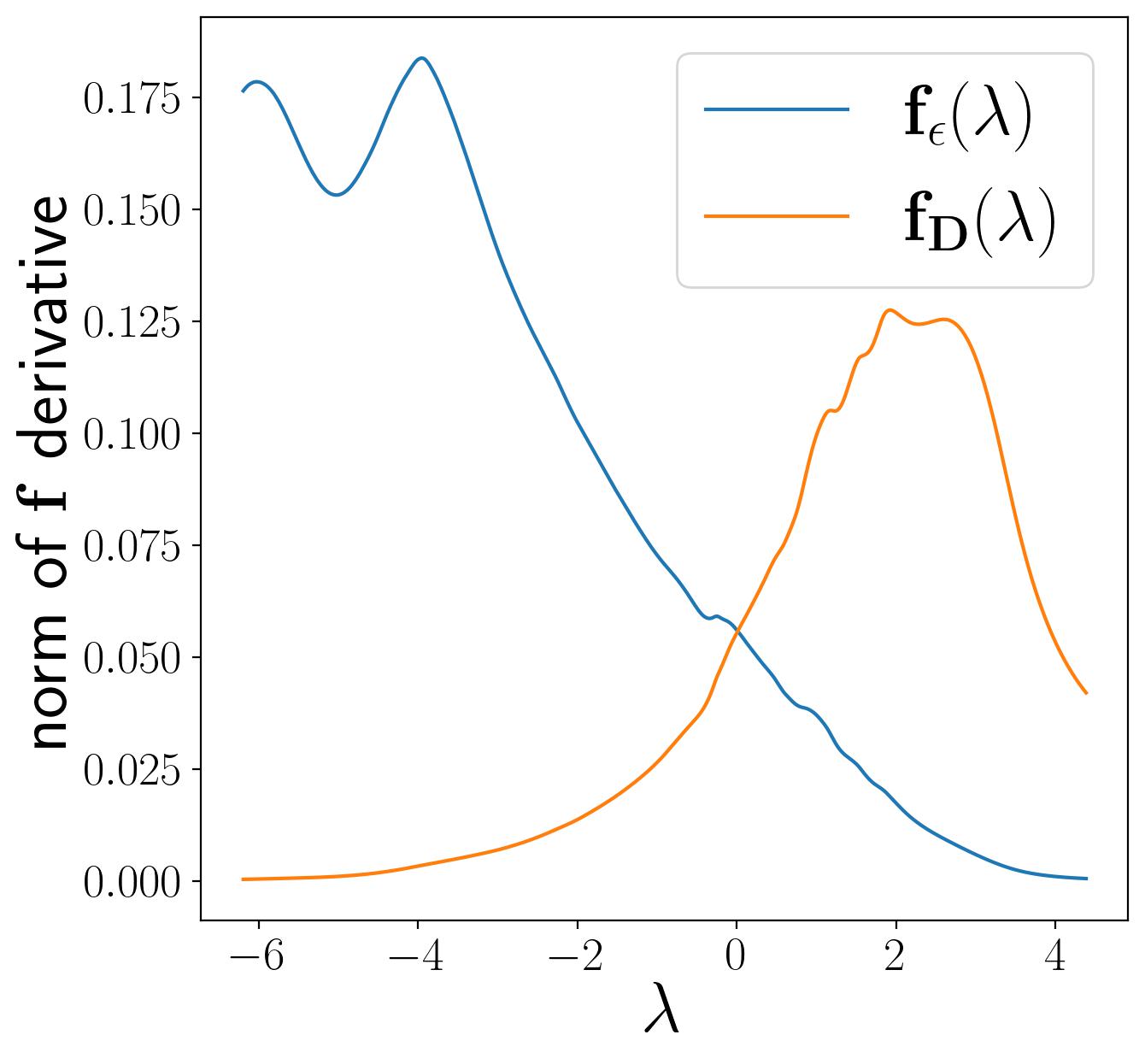}
    \caption{(Left) The evolution of nonlinear function evaluation for \cref{eq:eps-ode} characterized by $f_\epsilon (\sigma) := \epsilon_\theta(\vx(\sigma), \sigma)$.
(Right) The evolution of nonlinear function evaluation for \cref{eq:semi_ode_eps}, defined by $f_\epsilon(\lambda) := \epsilon_\theta(\vx(\lambda), \lambda)$ and $f_D(\lambda) := D_\theta(\vx(\lambda), \lambda)$.
Implementing a logarithmic transformation on the noise level $\sigma$ results in smoother trajectories of nonlinear function evaluations along ODE solutions.}
    \label{fig:app-function-derivative}
\end{figure}

\paragraph{Extra experiments} Besides the ``acceleration'' outlined in~\cref{fig:function-derivative}, we examine the evolution of $f(\lambda):=g(\vx(\lambda), \lambda)$ along the solution trajectory for various nonlinear functions $g$. We note that smoother $f$ typically results in smaller numerical defects, as indicated in~\Cref{lemma:intermediate-defect} and \cref{thm:order1,thm:order2}. As depicted in~\cref{fig:app-function-derivative}, the benefit of applying a logarithmic transformation to the noise level $\sigma$ is evident.

\subsection{Reproduction of~\cref{fig:numerical_imagenet}}

We utilize the pre-trained ImageNet $64\times 64$ class-conditioned diffusion model~\citep{karras2022elucidating} for our experiment. For the numerical defect experiments, we approximate the ground truth solution $\vx(0)$ using 500 steps RK4~(2000 NFE). We generate $50,000$ images with randomized labels to calculate numerical defects and FIDs. The EDM, as suggested by~\citet{karras2022elucidating}, follows a time schedule as given in~\cref{eq:time-rho}, 
\begin{equation}~\label{eq:time-rho}
    t_{i} = (
    \sigma_{\text{max}}^{\frac{1}{\rho}}
    + \frac{i}{N-1} (
        \sigma_{\text{min}}^{\frac{1}{\rho}} - \sigma_{\text{max}}^{\frac{1}{\rho}}
    )
    )^\rho,
\end{equation}
where $\rho$ is a hyperparameter that controls the timestamp spacing. Our findings reveal that the $\epsilon_\theta$-based \our outperforms the $\mD_\theta$-based \our.
By default, we adhere to the recommended $\rho = 7$ for our experiments. To test the robustness through the suboptimal time rescheduling experiment, we use $\rho=1$, where the timestamps are uniformly distributed in $\sigma$. For all experiments, we test the sampling algorithm with NFE set to $6, 8, 10, 15, 20, 25, 30, 35, 50, 75, 100$. Similar to EDM, an additional denoising step is included in the final stage. We note that for second-order single-step methods, the total NFE may not align perfectly with the NFE. In such instances, we always take one less step, for example, second-order methods only use 5 NFEs when we expect them to utilize 6 NFEs.
We use the official code from~\citet{karras2022elucidating} to calculate FID score. We include quantitative results in~\cref{tab:numerical_imagenet_left,tab:numerical_imagenet_mid,tab:numerical_imagenet_right}.

\begin{table}[]
    \centering
    \scalebox{0.9}{
    \begin{tabular}{c|cccccc}
    \toprule
        NFE& DDIM (S)& Heun (S)& DPM-Solver++ (S)& Our (S)& DPM-Solver++ (M)& Our (M)\\
        \midrule
6& 2515& 1.446$\times 10^4$$^{*}$& 2681$^{*}$& 1774$^{*}$& 1507& 2602\\
8& 2063& 3943$^{*}$& 2060$^{*}$& 1072$^{*}$& 1062& 1895\\
10& 1747& 2501$^{*}$& 1544$^{*}$& 729.1$^{*}$& 740.2& 1195\\
15& 1280& 992.7& 845.7& 382.6& 369.9& 463.3\\
20& 1020& 623.9$^{*}$& 617.8$^{*}$& 283.6$^{*}$& 227.6& 258.7\\
25& 852.7& 360.4& 420.7& 187.6& 153.2& 167.8\\
30& 731.4& 270.4$^{*}$& 339.4$^{*}$& 151.1$^{*}$& 114& 117.1\\
35& 640.1& 190.2& 256.5& 113.8& 87.17& 85.13\\
50& 468.4& 106.3$^{*}$& 146$^{*}$& 64.43$^{*}$& 46.06& 40.59\\
75& 328.1& 43.23& 68.56& 31.35& 22.82& 18.28\\
100& 254.6& 24.56$^{*}$& 41.63$^{*}$& 19.08$^{*}$& 14.58& 10.65\\
\bottomrule
    \end{tabular}
    }
    \caption{Numerical defects $|\vx(0) - \vx_N|$ with different NFEs and recommended time scheduling for~\cref{fig:numerical_imagenet}~(Left). $*$ indicates the number is produced by one less NFE.}
    \label{tab:numerical_imagenet_left}
\end{table}

\begin{table}[]
    \centering
    \scalebox{0.9}{
    \begin{tabular}{c|cccccc}
    \toprule
        NFE& DDIM (S)& Heun (S)& DPM-Solver++ (S)& Our (S)& DPM-Solver++ (M)& Our (M)\\
        \midrule
6& 41.09& 248.3$^{*}$& 68.97$^{*}$& 56.46$^{*}$& 14.87& 14.32\\
8& 25.12& 86.88$^{*}$& 28.33$^{*}$& 19.4$^{*}$& 8.024& 7.44\\
10& 17.26& 35.32$^{*}$& 16.77$^{*}$& 8.665$^{*}$& 5.456& 5.115\\
15& 9.284& 5.544& 6.402& 3.944& 3.525& 3.23\\
20& 6.446& 3.682$^{*}$& 4.593$^{*}$& 3.298$^{*}$& 2.975& 2.542\\
25& 5.101& 2.899& 3.493& 2.885& 2.726& 2.412\\
30& 4.348& 2.705$^{*}$& 3.151$^{*}$& 2.742$^{*}$& 2.602& 2.377\\
35& 3.712& 2.571& 2.86& 2.607& 2.523& 2.346\\
50& 3.18& 2.431$^{*}$& 2.575$^{*}$& 2.471$^{*}$& 2.425& 2.336\\
75& 2.968& 2.406& 2.514& 2.436& 2.401& 2.335\\
100& 2.755& 2.381& 2.453& 2.401& 2.377& 2.333\\
\bottomrule
    \end{tabular}
    }
    \caption{FID with different NFEs and recommended time scheduling for~\cref{fig:numerical_imagenet}~(Mid). $*$ indicates the number is produced by one less NFE.}
    \label{tab:numerical_imagenet_mid}
\end{table}

\begin{table}[]
    \centering
    \scalebox{0.9}{
    \begin{tabular}{c|cccccc}
    \toprule
        NFE& DDIM (S)& Heun (S)& DPM-Solver++ (S)& Our (S)& DPM-Solver++ (M)& Our (M)\\
        \midrule
6& 2339& 4.116$\times 10^4$$^{*}$& 2850$^{*}$& 1957$^{*}$& 5034& 4615\\
8& 1948& 1.249$\times 10^4$$^{*}$& 2004$^{*}$& 1177$^{*}$& 5281& 5430\\
10& 1664& 5901$^{*}$& 1578$^{*}$& 827.9$^{*}$& 5296& 5800\\
15& 1241& 1362& 907.2& 435.6& 4975& 5469\\
20& 996.9& 780.2$^{*}$& 681$^{*}$& 314.8$^{*}$& 4445& 4779\\
25& 834.9& 428& 483.1& 222.6& 3923& 4116\\
30& 718.5& 312.7$^{*}$& 392.9$^{*}$& 184.8$^{*}$& 3513& 3675\\
35& 632.6& 218.3& 300.5& 139.6& 3191& 3409\\
50& 467.5& 126.1$^{*}$& 185.6$^{*}$& 90.87$^{*}$& 2597& 2607\\
75& 329.4& 79.22& 101.2& 55.01& 2019& 1876\\
100& 256.8& 63.15$^{*}$& 70.82$^{*}$& 43.44$^{*}$& 1658& 1537\\
\bottomrule
    \end{tabular}
    }
    \caption{Numerical defects $|\vx(0) - \vx_N|$ with different NFEs and suboptimal time scheduling for~\cref{fig:numerical_imagenet}~(Right). $*$ indicates the number is produced by one less NFE.}
    \label{tab:numerical_imagenet_right}
\end{table}

\subsection{Reproduction of~\cref{fig:openai_guided_dm}}

We utilize pre-trained diffusion models equipped with a noise data classifier from~\citet{dhariwal2021diffusion}. These diffusion models are trained on discrete time analogous to DDPM. By employing an additional noise data classifier $p(\vc | \vx)$, we can derive a new diffusion model by incorporating the gradient of the classifier into the diffusion model, resulting in the following denoiser:
\begin{align}
    \hat{\mD}(\vx, \sigma | \vc) &= \vx + \sigma^2 \nabla_{\vx} \log p(\vx|\vc; \sigma)  \\
    & = \vx + \sigma^2 \nabla_{\vx} \log p(\vx ; \sigma) + \sigma^2 \nabla_{\vx} \log p(\vc |\vx ; \sigma),
\end{align}
where $p(\vc | \vx; \sigma)$ denotes a classifier for the label $\vc$ on noise data. In practice, researchers have observed that increasing the weight on the classifier can enhance performance, as expressed by:
\begin{align}
    \hat{\mD}(\vx, \sigma | \vc)  \approx \hat{\mD}_\theta(\vx, \sigma | \vc) + \omega \sigma^2 \nabla_{\vx} \log p(\vy |\vx ; \sigma),
\end{align}
where $\omega$ can be adjusted to values greater than $1$. In our experiment, we utilize the pre-trained classifier detailed in~\citet{dhariwal2021diffusion}. We approximate the ground truth $\vx(0)$ using 500 steps of the RK4 method. In addition, we apply dynamic thresholding~\citep{saharia2022photorealistic} to all methods evaluated in this experiment to ensure a fair comparison. We conduct the sampling tests with NFE values of $6, 7, 8, 9, 10, 12, 15, 18, 20, 30, 50$. It should be noted that for guided diffusion models, our approach based on $\mD_\theta$ proves superior.
We use the official code from~\citet{dhariwal2021diffusion} to calculate FID score.

\subsection{Reproduction of~\cref{fig:exp_stochastic_sampler}}

This experiment is designed to demonstrate how an enhanced single-step sampler can augment stochastic samplers' performance. We utilize the pre-trained ImageNet $64\times 64$ class-conditioned diffusion model~\citep{karras2022elucidating} for this purpose. The time schedule adheres to the recommended setting with $\rho=7$. With a fixed level of stochasticity, parameterized by $\eta$, we contrast the second order Heun method with our method at NFE values of $6, 8, 10, 15, 20, 25, 30, 35, 50, 75, 100, 150, 200, 350, 500$. We also examine the influence of $\eta$ on the FID with a fixed NFE of 75, iterating $\eta=0, 0.01, 0.025, 0.05, 0.15, 0.20, 0.30, 0.35, 0.40, 0.45, 0.50$. We observe that adding random noise can enhance perceptual quality in terms of FID when compared to deterministic sampling, although an excess of stochasticity can impair sampling quality. We use the official code from~\citet{karras2022elucidating} to calculate FID score.

\subsection{Reproduction of~\cref{fig:dfif}}

This experiment is designed to corroborate the effectiveness of our enhanced order analysis and \our in large-scale text-to-image diffusion models~\cite{deepfloyd23}, primarily against the benchmark set by DPM-Solver++~\citep{lu2022dpmp}. The initial experiments examine FID-CLIP scores under varying classifier-free guidance weights~\cite{ho2022classifier}. 
With classifier-free guidance, we can build a new diffusion model based on a conditional diffusion model and an unconditional model. Specifically, for either the data prediction model or the noise prediction model, denoted by their network as $g$, the newly proposed diffusion model adheres to the following protocol:
\begin{equation}
    \hat{g}(\vx, \sigma, \vc) := g(\vx, \sigma, \emptyset) + \omega (g(\vx, \sigma, \vc) - g(\vx, \sigma, \emptyset)),
\end{equation}
where$\omega$ serves as the guidance weight, $\emptyset$ represents the unconditional signal, and $\vc$ stands for the conditional signal\footnote{We slightly modify the classifier-free equation to align it with the implementation~\citep{deepfloyd23} available at~\url{https://github.com/deep-floyd/IF/blob/2ad4dff7cde3cce91f237270a3cc81cae4578015/deepfloyd_if/modules/base.py\#L114}}. 
We vary the guidance between $1,1.5,3,5,10$. We employ ViT-g-14~\cite{ilharco_gabriel_2021_5143773} for the calculation of the CLIP score and \hyperlink{https://github.com/mseitzer/pytorch-fid}{pytorch-fid} for FID. This experiment is carried out on the MS-COCO validation dataset. Owing to computational resource constraints, the experiment is performed at a resolution of $64 \times 64$ with a fixed 25 NFE. It should be noted that for this experiment, we merely substitute the degenerated Butcher tableau of DPM-Solver++ with our Butcher tableau, keeping all other configurations consistent. The experiment is based on \hyperlink{https://huggingface.co/DeepFloyd/IF-I-XL-v1.0}{IF-I-XL}.

In a further experiment, we compare \our and DPM-Solver++ using the upsampling diffusion model \hyperlink{https://huggingface.co/DeepFloyd/IF-II-L-v1.0}{IF-II-L}, which upsamples a low-resolution image of $64\times 64$ pixels to a high-resolution image of $256 \times 256$ pixels. For this test, we feed generated $64 \times 64$ images, using the single-step \our with $49$ NFE, along with corresponding captions into the upsampling model. We then vary the NFE to assess the quality of the generated images. Intriguingly, our findings reveal that the single-step \our outperforms multi-step methods in terms of the quality of samples produced.

In our experiments involving DeepFloyd IF models, we adopt the time schedule outlined in~\cref{eq:time-rho} with $\rho =7$. Owing to constraints on our computational resources, we are unable to sweep through hyperparameters to locate a more optimized time schedule. However, we posit that with the same NFE, the quality of the sampling could be further enhanced by fine-tuning the time schedule.

\subsection{Licenses}

\paragraph{Dataset}

\begin{itemize}
    \item \makebox[3cm][l]{ImageNet~\cite{russakovsky2015imagenet}} The license status is unclear
    \item \makebox[3cm][l]{MS-COCO~\cite{lin2014microsoft}} Creative Commons Attribution 4.0 License.
\end{itemize}

\paragraph{Pre-trained models and code}

\begin{itemize}
    \item \makebox[8cm][l]{EDM ImageNet \citep{karras2022elucidating}}   Creative Commons Attribution-NonCommercial-ShareAlike 4.0 International License
    \item \makebox[8cm][l]{DeepFloyd IF~\citep{deepfloyd23}} DeepFloyd IF License Agreement
    \item \makebox[8cm][l]{OpenAI Guided ImageNet ~\citep{dhariwal2021diffusion}} MIT License
    \item \makebox[8cm][l]{Pytorch FID~\cite{Seitzer2020FID}} Apache License 2.0
    \item \makebox[8cm][l]{OpenCLIP~\cite{ilharco_gabriel_2021_5143773}} MIT License
\end{itemize}
\section{Image samples}

In this section, we include samples from our experiments for quantitative comparison.

\begin{figure}
    \centering
    \includegraphics[width=0.8\textwidth]{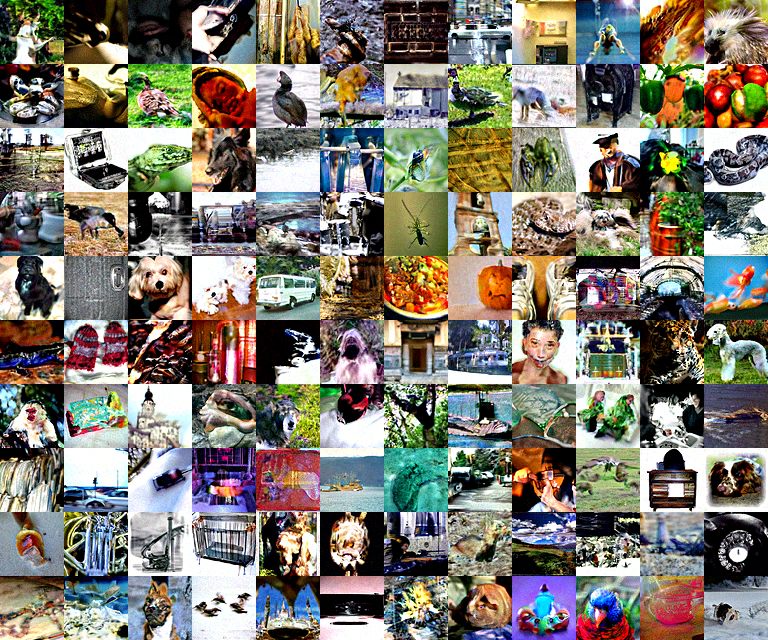}
    \caption{Uncurated samples of $64\times 64$ ImageNet model~\cite{karras2022elucidating} with single-step Heun 9 NFE. (FID=35.31)}
\end{figure}
\begin{figure}
    \centering
    \includegraphics[width=0.8\textwidth]{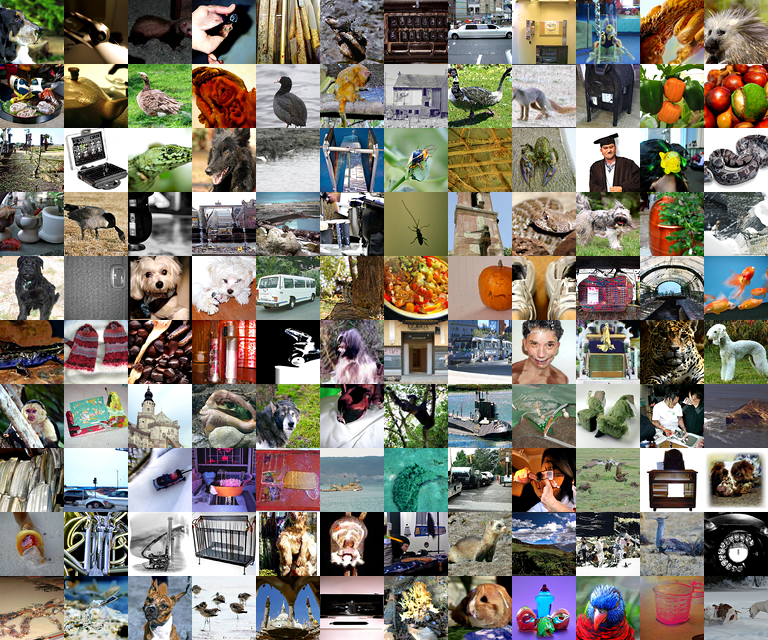}
    \caption{Uncurated samples of $64\times 64$ ImageNet model~\cite{karras2022elucidating} with single-step Heun 15 NFE. (FID=5.54)}
\end{figure}

\begin{figure}
    \centering
    \includegraphics[width=0.8\textwidth]{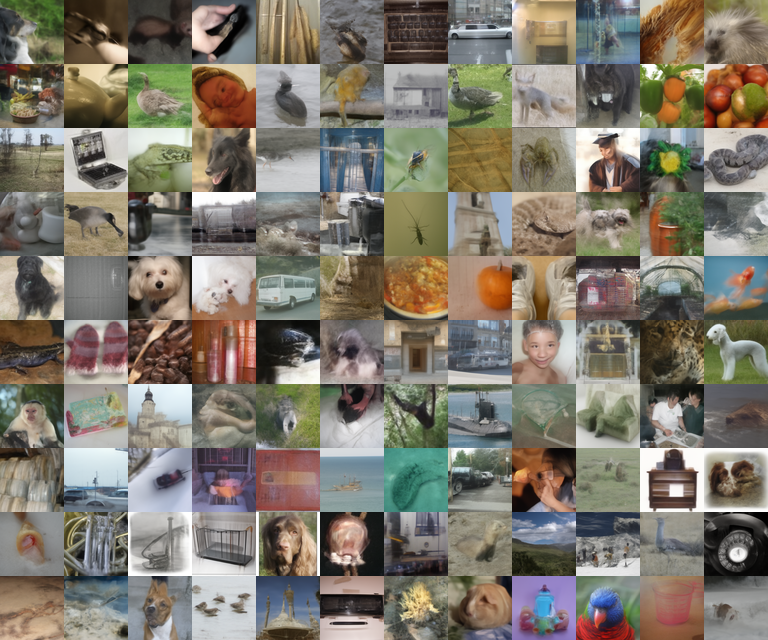}
    \caption{Uncurated samples of $64\times 64$ ImageNet model~\cite{karras2022elucidating} with single-step DPM-Solver++ 9 NFE. (FID=16.77)}
\end{figure}
\begin{figure}
    \centering
    \includegraphics[width=0.8\textwidth]{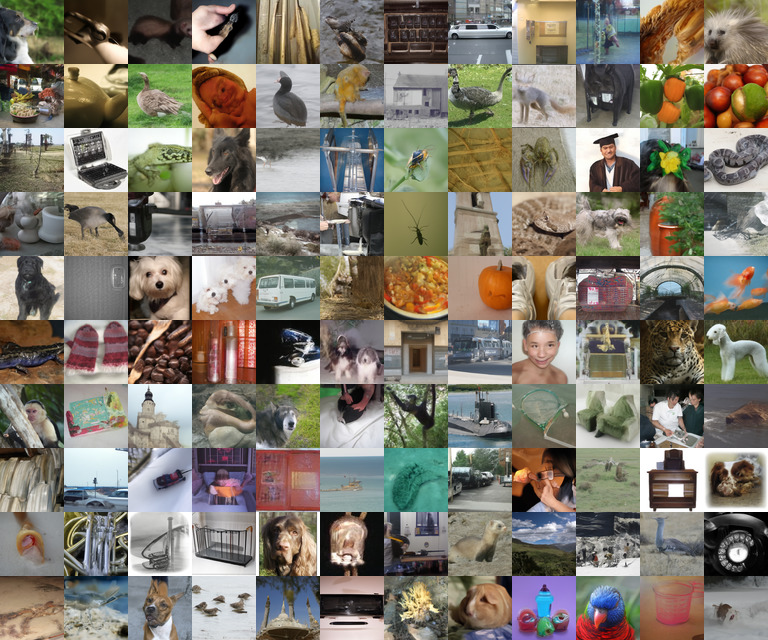}
    \caption{Uncurated samples of $64\times 64$ ImageNet model~\cite{karras2022elucidating} with single-step DPM-Solver++ 15 NFE. (FID=6.40)}
\end{figure}

\begin{figure}
    \centering
    \includegraphics[width=0.8\textwidth]{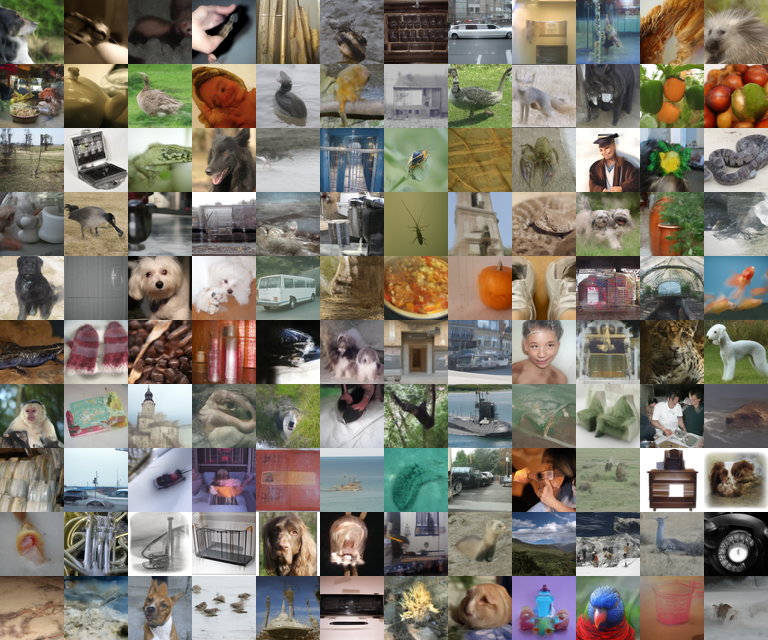}
    \caption{Uncurated samples of $64\times 64$ ImageNet model~\cite{karras2022elucidating} with single-step \our data prediction model 9 NFE. (FID=8.66)}
\end{figure}
\begin{figure}
    \centering
    \includegraphics[width=0.8\textwidth]{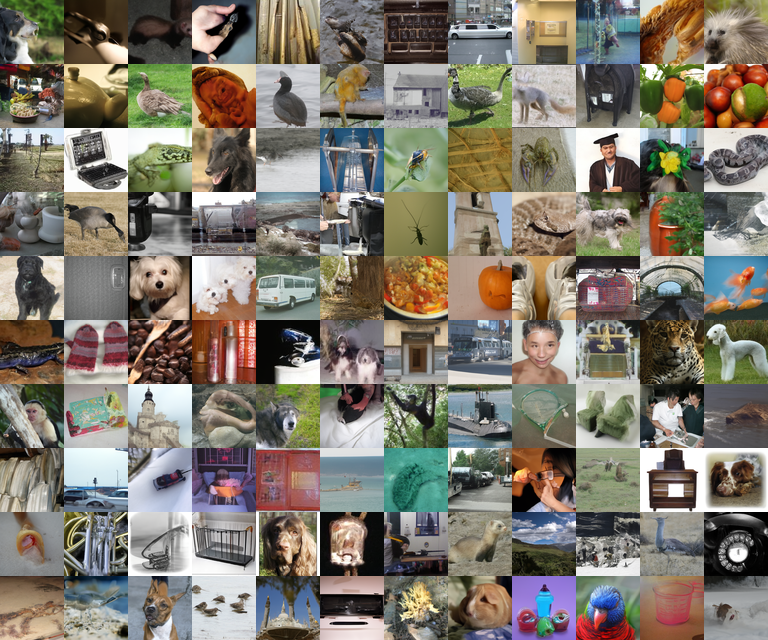}
    \caption{Uncurated samples of $64\times 64$ ImageNet model~\cite{karras2022elucidating} with single-step \our data prediction model 15 NFE. (FID=3.92)}
\end{figure}

\begin{figure}
    \centering
    \includegraphics[width=0.8\textwidth]{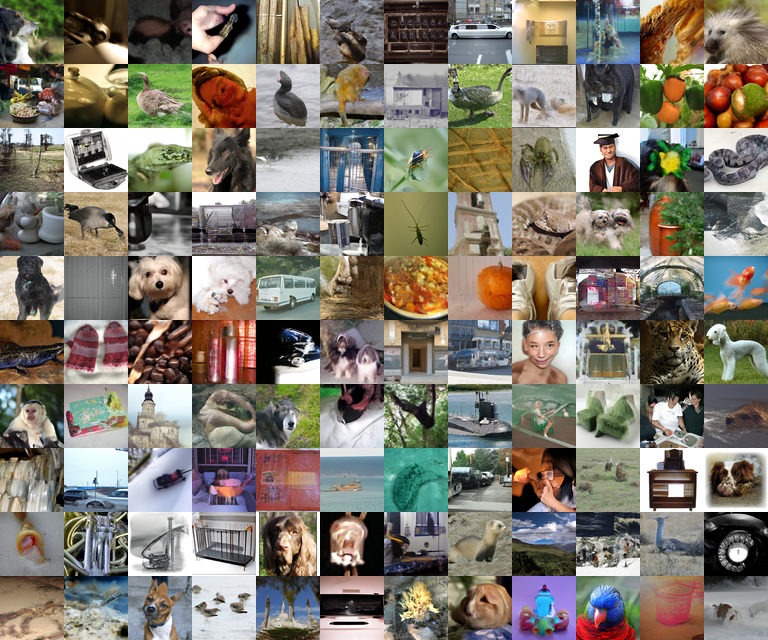}
    \caption{Uncurated samples of $64\times 64$ ImageNet model~\cite{karras2022elucidating} with single-step \our noise prediction model 9 NFE. (FID=5.08)}
\end{figure}
\begin{figure}
    \centering
    \includegraphics[width=0.8\textwidth]{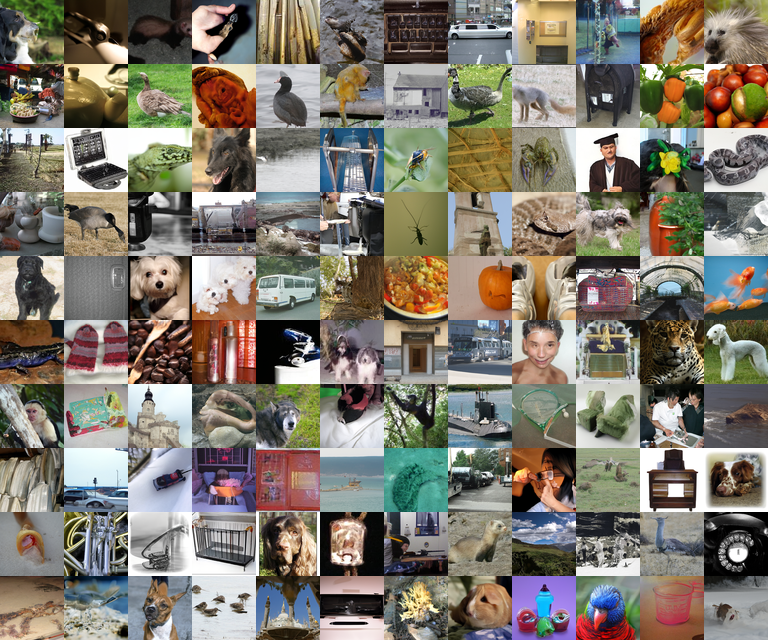}
    \caption{Uncurated samples of $64\times 64$ ImageNet model~\cite{karras2022elucidating} with single-step \our noise prediction model 15 NFE. (FID=3.88)}
\end{figure}

\begin{figure}
    \centering
    \includegraphics[width=\textwidth]{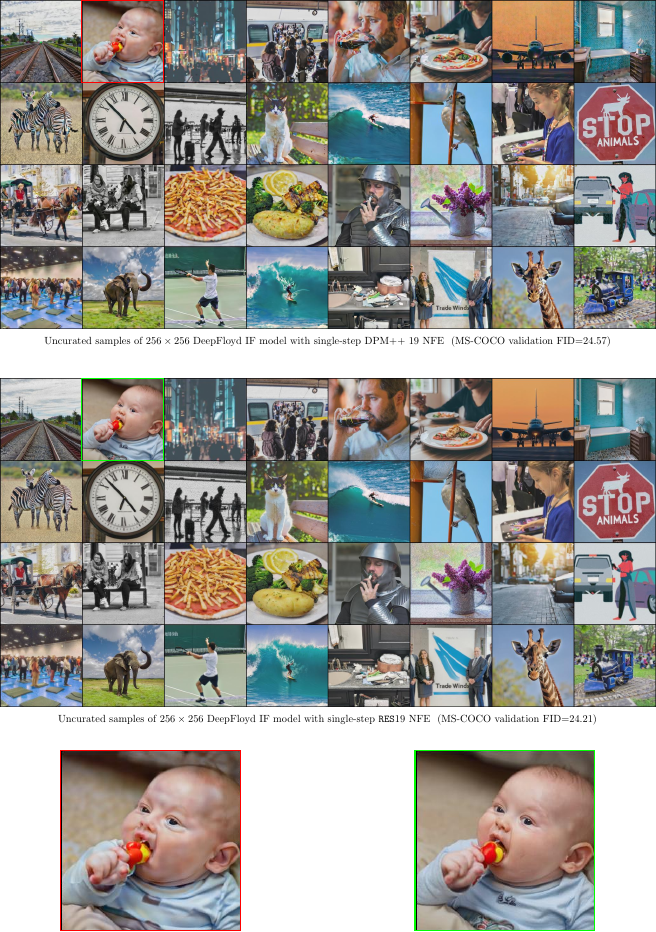}
    \caption{ Single-step DPM-Solver++  \textit{vs} Single-step \our on cascaded DM}
    \label{fig:app-dfif}
\end{figure}

\end{document}